\documentclass{article}

\PassOptionsToPackage{numbers, compress}{natbib}
\usepackage[preprint]{neurips_2025}
\usepackage[utf8]{inputenc} % allow utf-8 input
\usepackage[T1]{fontenc}    % use 8-bit T1 fonts
\usepackage{hyperref}       % hyperlinks
\usepackage{url}            % simple URL typesetting
\usepackage{booktabs}       % professional-quality tables
\usepackage{amsfonts}       % blackboard math symbols
\usepackage{nicefrac}       % compact symbols for 1/2, etc.
\usepackage{microtype}      % microtypography
\usepackage{xcolor}         % colors

\usepackage{multirow}
\usepackage{amssymb, amsmath}
\usepackage{siunitx}
\usepackage{bm}
\usepackage{xfp}
\usepackage{comment}
\usepackage{xspace}
\usepackage{accents}
\usepackage{animate}
\usepackage{subcaption} 
\usepackage{colortbl}
\usepackage{rotating}
\usepackage[accsupp]{axessibility}  % Improves PDF readability for those with disabilities.
\usepackage{pifont}% http://ctan.org/pkg/pifont
\usepackage{array}  
\usepackage{geometry}
\usepackage{tcolorbox}
\usepackage{lipsum}
\usepackage{makecell}
\usepackage{enumitem}
\usepackage{algpseudocode}
\usepackage{algorithm}

\usepackage{wrapfig}

\usepackage[capitalize]{cleveref}
\crefname{section}{Sec.}{Secs.}
\Crefname{section}{Section}{Sections}
\Crefname{table}{Table}{Tables}
\crefname{table}{Tab.}{Tabs.}
\definecolor{maroon}{cmyk}{0,0.87,0.68,0.32}
\definecolor{myyellow}{RGB}{218, 160, 109}
\definecolor{brickred}{rgb}{0.8, 0.25, 0.33}
\definecolor{brandeisblue}{rgb}{0.0, 0.44, 1.0}
\definecolor{applegreen}{rgb}{0.55, 0.71, 0.0}
\definecolor{aogreen}{rgb}{0.0, 0.5, 0.0}
\definecolor{turquoise}{cmyk}{0.65,0,0.1,0.3}
\definecolor{purple}{rgb}{0.65,0,0.65}
\definecolor{dark_green}{rgb}{0, 0.5, 0}
\definecolor{orange}{rgb}{0.8, 0.6, 0.2}
\definecolor{red}{rgb}{0.8, 0.2, 0.2}
\definecolor{darkred}{rgb}{0.6, 0.1, 0.05}
\definecolor{blueish}{rgb}{0.0, 0.3, .6}
\definecolor{light_gray}{rgb}{0.7, 0.7, .7}
\definecolor{pink}{rgb}{1, 0, 1}
\definecolor{greyblue}{rgb}{0.25, 0.25, 1}
\definecolor{orgred}{rgb}{1.0, 0, 0}
\usepackage{pifont}
\newcommand{\cA}{\mathcal{A}}

\newcommand{\cD}{\mathcal{D}}

\newcommand{\cX}{\mathcal{X}}

\usepackage{scalerel,stackengine}
\stackMath
\newcommand\reallywidehat[1]{%
\savestack{\tmpbox}{\stretchto{%
  \scaleto{%
    \scalerel*[\widthof{\ensuremath{#1}}]{\kern-.6pt\bigwedge\kern-.6pt}%
    {\rule[-\textheight/2]{1ex}{\textheight}}%WIDTH-LIMITED BIG WEDGE
  }{\textheight}% 
}{0.5ex}}%
\stackon[1pt]{#1}{\tmpbox}%
}
\newcommand\reallywidecheck[1]{%
\savestack{\tmpbox}{\stretchto{%
  \scaleto{%zh
    \scalerel*[\widthof{\ensuremath{#1}}]{\kern-.6pt\bigwedge\kern-.6pt}%
    {\rule[-\textheight/2]{1ex}{\textheight}}%WIDTH-LIMITED BIG WEDGE
  }{\textheight}% 
}{0.5ex}}%
\stackon[1pt]{#1}{\scalebox{-1}{\tmpbox}}%
}

\title{{MoDoMoDo}: \underline{Mul}ti-\underline{Do}main Data Mixtures\\for \underline{Mul}timo\underline{dal} LLM Reinforcement Learning}

\author{%
Yiqing Liang$^{1}$\footnotemark[1],  Jielin Qiu$^2$, Wenhao Ding$^3$, Zuxin Liu$^2$, James Tompkin$^1$,\\ 
\bf Mengdi Xu$^4$, Mengzhou Xia$^5$, Zhengzhong Tu$^6$, Laixi Shi$^7$, Jiacheng Zhu$^8$\footnotemark[1]\\[2pt]
$^1$Brown University\quad 
$^2$Salesforce AI Research \quad
$^3$NVIDIA Research \\
$^4$Carnegie Mellon University \quad
$^5$Princeton University \quad
$^6$Texas A\&M University \\
$^7$California Institute of Technology \quad
$^8$MIT CSAIL\quad
\\
\\[6pt]
\textbf{\textcolor{magenta}{Project Website}}: \url{https://modomodo-rl.github.io/}
}

\begin{document}

\maketitle

\footnotetext[1]{* corresponding: yiqing\_liang@brown.edu, zjc@mit.edu}

\begin{abstract}
Reinforcement Learning with Verifiable Rewards (RLVR) has recently emerged as a powerful paradigm for post-training large language models (LLMs), achieving state-of-the-art performance on tasks with structured, verifiable answers. 
Applying RLVR to Multimodal LLMs (MLLMs) presents significant opportunities but is complicated by the broader, heterogeneous nature of vision-language tasks that demand nuanced visual, logical, and spatial capabilities. 
As such, training MLLMs using RLVR on multiple datasets could be beneficial but creates challenges with conflicting objectives from interaction among diverse datasets, highlighting the need for optimal dataset mixture strategies to improve generalization and reasoning.
We introduce a systematic post-training framework for Multimodal LLM RLVR, featuring a rigorous data mixture problem formulation and benchmark implementation. 
Specifically, 
(1) We developed a multimodal RLVR framework for multi-dataset post-training by curating a dataset that contains different verifiable vision-language problems and enabling multi-domain online RL learning with different verifiable rewards; 
(2) We proposed a data mixture strategy that learns to predict the RL fine-tuning outcome from the data mixture distribution, and consequently optimizes the best mixture. 
Comprehensive experiments showcase that multi-domain RLVR training, when combined with mixture prediction strategies, can significantly boost MLLM general reasoning capacities. 
Our best mixture improves the post-trained model's accuracy on out-of-distribution benchmarks by an average of $5.24\%$ compared to the same model post-trained with uniform data mixture, and by a total of $20.74\%$ compared to the pre-finetuning baseline.
%Our best mixture brings an average of $5\%$ accuracy increase to the post-trained model on out-of-distribution benchmarks compared to uniform sampling, and a total of $15.5\%$ accuracy compared to before finetuning. 
%{\color{red}which demonstrate [XXX quantitative superior performance] compared to using a single dataset/or some baselines}
 
% Our contributions include: (1) Developing a multi-objective RLVR post-training pipeline that incorporates dataset mixture optimization and comprehensive in-domain/out-of-domain evaluations; (2) Introducing a data-driven method to predict optimal dataset mixing weights based on pilot training runs, thus balancing performance across varied vision-language tasks; and (3) Analyzing the theoretical foundations of multi-objective reinforcement learning for multimodal scenarios. Empirical evaluations demonstrate that our optimized data mixing strategy effectively generalizes across diverse multimodal tasks, achieving robust performance while mitigating issues such as reward hacking and overfitting.
\end{abstract}

% \mz{I like the motivation in the abstract. But the exposition leading up to it feels a bit lengthy and could be more direct and concise.}

% \mz{The contributions could be clarified further. Specifically, how are (1) and (2) different? (1) seems to describe a pipeline, while (2) is presented as a method—does (1) subsume (2)? If so, it may help to make that relationship explicit, or make each point mutually exclusive. For (3), could we make the claim more quantitative? For example, stating that method XXX improves baseline performance by YYY points would strengthen the contribution.}

% % \lxs{revised according to mengzhou's wonderful suggestions, but leave the final sentence for yiqing/jiacheng since it need detailed performance evidence.}

% \yl{write at the tail of abstract that we would release all code, data?}

\begin{figure*}
\includegraphics[width=\textwidth]{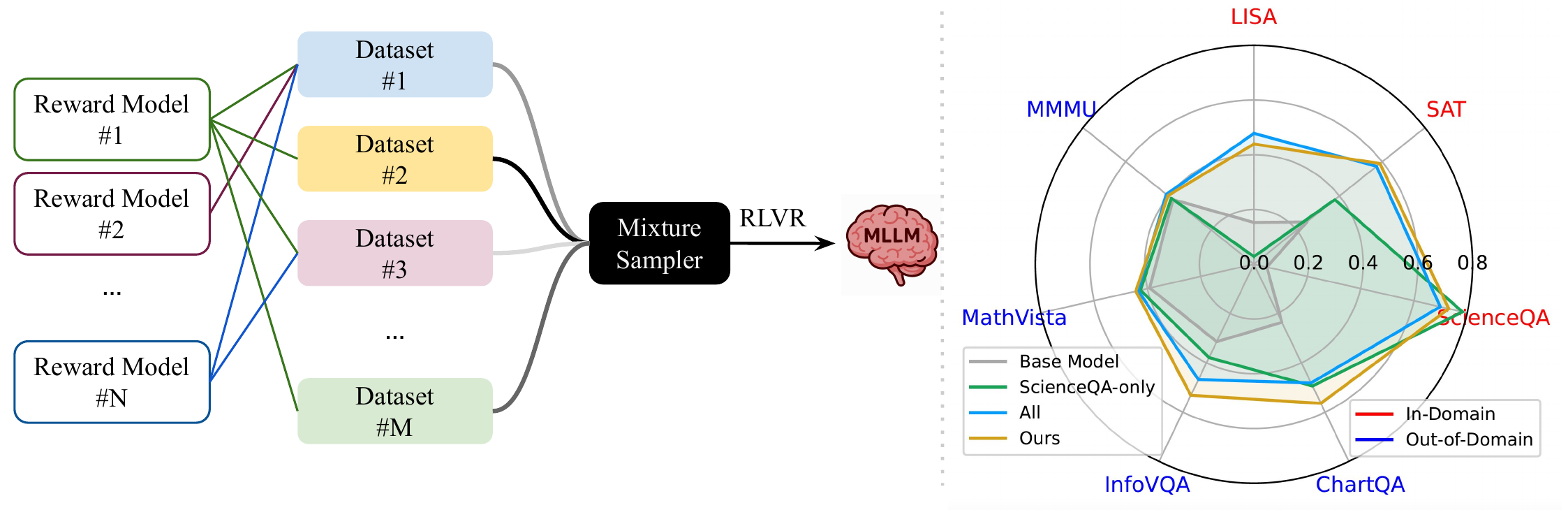}
    %\vspace{-10pt}
    \captionof{figure}{\textbf{MoDoMoDo} is a framework that combines \emph{Multi-Domain Data Mixtures} with \emph{Multimodal LLM Reinforcement Learning}, enabling generalizable performance gain across diverse VL tasks. Models trained with our estimated optimal mixtures can outperform those trained with naive mixtures on in-domain and out-of-domain benchmarks.}
    \vspace{-5mm}
    % }
    \label{fig:teaser}
\end{figure*}

\section{Introduction}

% \tzz{I feel it's better to start the introduction with RL. Then say RL-based training is popular for LLMs. Then introduce PPO, GRPO, RLVR, etc. }
% \tzz{Par2: mention the current status quo of RL for MLLM, mention a few representative methods. However, gaps and challenges, that motivates our work}
% \tzz{Next is the summary of this work + contributions.}

Multimodal large language models (MLLMs)\cite{openai2023gpt4} are significantly expanding the frontiers of artificial intelligence. They integrate diverse data types like images and audio with text to achieve a more comprehensive understanding and enable richer, human-natural AI interactions, unlocking advanced capabilities such as cross-modal reasoning and spatial intelligence \cite{radford2021learning,alayrac2022flamingo,liu2023visual,mckinzie2024mm1}. 
Effective post-training of multimodal LLMs remains challenging as they must integrate diverse data modalities, whose data is orders of magnitude scarcer than text-only LLMs~\cite{sun2023aligning_vision} and present fundamentally different problem features. 
They translate into a broad spectrum of distinct capabilities—spatial reasoning, fine‑grained recognition, chart interpretation, and more.
% \yl{These data translate into a broad spectrum of capabilities—spatial reasoning, fine‑grained recognition, chart interpretation, and more}. 
In response, researchers have developed and adapted various fine-tuning techniques for MLLMs, including instruction following~\cite{liu2023visual_instruction_tuning_llava}, Reinforcement Learning from Human Feedback (RLHF) \cite{Ouyang2022instructgpt,sun2023aligning_vision}, Direct Preference Optimization (DPO)\cite{rafailov2023_dpo,wang2024_mdpo}, and online reinforcement learning (RL) approaches~\cite{Guo2025DeepSeekR1,shen2025vlmr1,chen2025r1v}.
% This challenge is compounded by the need for MLLMs to integrate diverse sensory modalities—such as vision, audio, and touch—to emulate the way humans perceive and interact with their environment. 

% \mz{I have a similar comment here as the abstract, the introduction part is a bit lengthy. The first paragraph outlines general fine-tuning techniques for MLLMs, the second introduces RLVR, and the third aims to describe the challenges of extending RLVR to MLLMs. However, it leans more toward discussing the desired capabilities of MLLMs rather than challenges specific to RLVR. I think the background and the setup of challenges in this introduction could be made crisper. I would suggest briefly summarizing the current state of post-training for MLLMs, then directly stating the core challenge: balancing diverse datasets in RL due to the wide range of capabilities we aim to develop.}

Recently, reinforcement learning with verifiable rewards (RLVR)\cite{deepseek-math,mroueh2025rlvr} has emerged as a powerful post-training paradigm for text-based language models, delivering state-of-the-art performance in domains with programmatically checkable answers, such as mathematics and code. By replacing noisy human preferences with rule-based signals, RLVR yields stable optimization and strong generalization\cite{deepseek-math,Guo2025DeepSeekR1,mroueh2025rlvr}. However, using RLVR to enable MLLMs presents significant challenges due to the relative scarcity of accessible verifiable datasets compared to the huge demand for data driven by inherently richer and heterogeneous multimodal tasks. Each verifiable reward tackles a slice of that capability spectrum, leaving large gaps in multimodal reasoning.
% \yl{While valuable, Each effort unlocks only a narrow slice of that capability spectrum, leaving large gaps in multimodal reasoning.} 
Existing effective attempts that transfer vision tasks into verifiable reward signals for RL fine-tuning often only focus on one specific multimodal task domain (e.g., VQA~\cite{chen2025r1v}, object detection~\cite{shen2025vlmr1}, math~\cite{skywork2025r1v,li2025openr1multimodal}, and video~\cite{feng2025videor1}). 
% \yl{
Such single‑domain focus is inadequate for achieving the wide range of capabilities required by MLLMs.
% }
%This focus can be inadequate for achieving the desirable generalization and comprehensive reasoning capabilities of MLLMs.
% We aim to answer the following question: How to balance diverse datasets in RL due to the wide range of capabilities we aim to develop.

To address this, it is beneficial to incorporate multiple datasets from diverse multimodal task domains during post-training, rather than relying on a single dataset, to ensure broad coverage and generalization across a wide range of capabilities. This is particularly important for multimodal LLMs, given the vast number of distinct tasks arising from multimodal compositions. However, using multiple training datasets introduces challenges, including potential conflicting objectives resulting from interactions among diverse datasets, as well as corresponding unstable behaviors during training. These challenges underscore the need to optimize dataset mixture strategies. Notably, data mixture has been extensively investigated and has demonstrated its effectiveness in various scenarios, such as supervised deep learning, pre-training LLMs, and supervised fine-tuning.
Here we seek to answer: 
% \textit{How to balance diverse datasets in RL due to the wide range of capabilities we aim to develop?}
\begin{tcolorbox}[before skip=1mm, after skip=0.0cm, boxsep=0.0cm, middle=0.0cm, top=0.1cm, bottom=0.1cm]
    \textit{\textbf{(Q)}
    How to mix diverse datasets in RLVR to achieve the wide range of multimodal capabilities?
    }
\end{tcolorbox}
\vspace*{0.2cm}

To enable efficient multi-dataset multimodal RLVR for multimodal LLM, our main contributions are:
\begin{itemize}[noitemsep,topsep=0pt,leftmargin=1.5em]
    \item We develop a \textbf{novel multimodal RLVR framework} for \textbf{multi-dataset post-training}, designed to optimize the strategies for mixing these datasets to achieve desired capabilities. 
    This includes the creation of five image-text datasets with different verifiable rewards, as well as the RLVR framework for training on these datasets~(\cref{sec:data mixture_implementation}).
    \item We propose a \textbf{multi-domain data mixture optimization strategy}, which learns a surrogate function to predict the outcome of RL finetuning given the mixture distribution~(\cref{sec:data mixture_modeling}). We show that a multivariate quadratic function can capture the counterfactual interplay among different datasets~(\cref{fig:min_max_var}).  
    \item  
    % By employing different data mixture strategies for these datasets during training, 
    We comprehensively \textbf{evaluate the performance on various evaluation sets}, including MMMU, MathVista, InfoVQA, and ChartQA. The predicted optimal data mixture distribution drives the finetuned model to have strong and generalizable reasoning capability, compared with any other standalone or naive mixture recipes~(\cref{sec:analysis}). 
    % The results demonstrate significant enhancements in generalization ability after incorporating multiple training datasets with appropriate mixture strategies.

    % We use a surrogate function to learn the scaling properties of datamix distributions.  
    % % that can achieve a balance between multiple ID/OOD performance and generalize well, 
    % where we estimate the optimal mixing weights from a small set of pilot runs
    % \item Understand this process through multi-objective reinforcement learning (theoretical results?)
    % \item We evaluate the xxxxx
\end{itemize}

\section{Multimodal RLVR with Data Mixtures}

% We first outline key definitions
% relevant to our analysis and provide an overview of the general experimental setup.

% Problem formulation of data mixture with RLVR
%% Multiple dataset
%% RLVR with such multiple dataset
% our approach for this problem
%% using a prior model for this problem and then using pre-train data to estimate the model and then calculte the optimal data mixtures

In this section, we first present the problem formulation of RLVR with data mixtures. Subsequently, we propose an approach to optimize the data mixture strategy using a two-step procedure, incorporating a prior model for the objective of the inner optimization problem.

\subsection{Reinforcement Learning with Verifiable Reward (RLVR) with Data Mixtures}

\paragraph{Background: Reinforcement Learning with Verifiable Reward (RLVR) for Post-Training.}
A large language model (LLM) is designed to process a prompt  $x\in\cX$ and generate a response $a\in\cA$ according to a policy $\pi: \cX \mapsto \Delta(\cA)$. The training of LLMs typically involves two phases: pre-training and post-training, both utilizing the same model architecture. After pre-training, a reference policy $\pi_{\mathsf{ref}}: \mathcal{X} \rightarrow \Delta(\mathcal{A})$ is obtained.

In the post-training phase, Reinforcement Learning with Verifiable Reward (RLVR) has emerged as a powerful and scalable approach to align LLMs with desired capabilities and preferences, such as reasoning. RLVR can be viewed as a variant of reinforcement learning with human feedback (RLHF), but instead of relying on a learned reward model from human preferences, it utilizes verifiable, criteria-based rewards $r: \mathcal{X} \times \mathcal{A} \rightarrow [0,1]$ that directly encode reasoning and accuracy requirements \cite{guo2025deepseek}. For instance, the reward can be (1) verifiable correctness, (2) verification via execution, or (3) verifiable constraints. These rewards are typically more transparent and less susceptible to reward hacking compared to learned reward models, which often require careful design and tuning for specific datasets.

Suppose we have access to a dataset of prompts $\mathcal{D} = \{x_i\}$ drawn from a distribution $x_i \sim \mu \in \Delta(\mathcal{X})$. Let the policy $\pi_\theta$ be parameterized by $\theta$. Except for the reward specification, RLVR for post-training follows the same pipeline as RLHF and is formulated as:
\begin{align}
    & \max_{\theta} L(\mu,\theta) := \max_{\theta} \mathbb{E}_{x\in \mu, a \sim \pi_{\theta}(\cdot | x)} \left[ r(x,a)- \beta D_{\text{KL}} (\pi_{\theta}(\cdot | x) \| \pi_{\mathsf{ref}}(\cdot | x)  \right], \label{eq:RLVR}
 \end{align}
where  $D_{\text{KL}}(\cdot)$ represents the Kullback–Leibler (KL) divergence.

\paragraph{RLVR with Data Mixtures.} 
In practice, we often aim to fine-tune an LLM to perform well across a variety of tasks, which naturally benefits from training on multiple datasets spanning diverse domains. Without loss of generality, suppose we have access to $m$ multimodal datasets $\cD = \{\mathcal{D}_i, r_i\}_{i=1}^m$, each associated with a domain-specific verifiable reward function $r_i: \mathcal{X} \times \mathcal{A} \rightarrow [0,1]$. Each dataset $\cD_i = \{ x_{i,j}\}_{j=1}^{|\cD_i|} $ contains prompts drawn from distribution $x_{i,j}\sim\mu_i$. The ultimate goal is to achieve an optimally generalizable MLLM in a testing prompt distribution $\mu_{\mathsf{test}}$ with corresponding verifiable reward $r_{\mathsf{test}}$, which potentially not only encompasses the tasks within the $m$ domains used during training but also extends to another $n$ out-of-distribution domains (multimodal tasks).

Now we are ready to provide the formulation of Multimodal RLVR with Data Mixtures. 
% Consider any LLM model with parameters $\theta$, and let the evaluation metric for eachbe defined as: $ \mathbb{E}_{x\in \mu_{\mathsf{test}}} L_{\mathsf{test}}(x, \theta)$, where $\mu_{\mathsf{test}}$ is the test distribution, potentially different from any training distribution and possibly including unseen prompts. 
We consider training the LLM using a mixture of the $m$ datasets, weighted by $w \in \Delta_{m}$, where $w$ is a probability distribution over the $m$ datasets. This results in a mixed training distribution: $\mu(w) = \sum_{i=1}^{m} w_i \mu_i$. The objective of RLVR with data mixtures is to optimize $w$ such that the resulting model performs best on the test distribution, i.e., 
\begin{align}
\max_{w}  L_{\mathsf{test}}(w) := \max_{w}  \mathbb{E}_{x\in \mu_{\mathsf{test}}} \big[\mathbb{E}_{a\sim \pi_{\theta(w)^\star}(\cdot | x)}[r_{\mathsf{test}}(x,a)] \big] , \quad \text{ s.t } \quad \theta(w)^\star = \arg\max_{\theta} L( \mu(w),\theta). \label{eq:RLVR-mixing}
\end{align}
Here, $\theta(w)^\star$ denotes the optimal model parameters obtained by running RLVR (as in Eq.~\eqref{eq:RLVR}) on the mixed dataset $\mu(w)$. In practice, this optimum may not be perfectly attainable, but the formulation can be generalized to any post-training procedure that outputs an LLM model with some parameters $\theta$ using the mixed distribution $\mu(w)$. 

\subsection{Model-Based Data Mixture Strategy}

To solve \eqref{eq:RLVR-mixing}, the objective is to find the optimal data mixture strategy $w$ for a fixed given distribution $\mu_{\mathsf{test}}$. Various techniques can be used to address this problem; however, a significant challenge lies in the fact that for a fixed $w$, its corresponding performance $\mathbb{E}_{x \in \mu_{\mathsf{test}}} \big[\mathbb{E}_{a \sim \pi{\theta(w)^\star}(\cdot | x)}[r(x,a)] \big]$ is unknown. Estimating this performance requires an entire post-training process (solving \eqref{eq:RLVR}), which makes searching through the exponentially large space $\Delta_m$ computationally infeasible. To overcome this challenge, we adopt a model-based approach \cite{ilyas2022datamodels}. Specifically, we first develop a model estimation $\widehat{L}_{\mathsf{test}}(w)$ for the ground truth performance function $L_{\mathsf{test}}(w)$ with relatively low computational cost. This estimation $\widehat{L}_{\mathsf{test}}(w)$ is then used to determine the optimal $w$ and the corresponding data mixture strategy $\mu(w)$ for RLVR post-training.

In particular, we use supervised learning to estimate the performance model $\widehat{L}_{\mathsf{test}}(w)$ and then compute its corresponding optimal data mixture strategy by following these key steps: 
\begin{itemize}
    \item \textbf{Construction of the training dataset for $\widehat{L}_{\mathsf{test}}(w)$}. First, construct a collection $\{ w^0_i \}_{i=1}^{k}$ as "seed distribution" samples. These seed distributions $\{ w^0_i \}_{i=1}^{k}$ are carefully chosen to maximize information while minimizing computational cost. This selection is inspired by the hypothesis that the different $m$ multimodal task domains might have independent or correlated effects on the $m+n$ testing tasks within $\mu_{\mathsf{test}}$. For example, we might choose $w^0_1 = [1, 0, 0, 0, 0]^\top$, $w^0_2 = [0, 1, 0, 0, 0]^\top$, and $w^0_3 = [0, 0.25, 0.25, 0.25, 0.25]^\top$ for $m = 5$. Then perform $k$ RLVR training runs to curate a collection of data $\{(w^0_1, L_{\mathsf{test}}(w^0_1)), (w^0_2, L_{\mathsf{test}}(w^0_2)), \dots \}_{i=1}^k$. We assume that the trained RLVR output approximates the optimal solution $\theta(w^0_i)$ for all $\{ w^0_i \}_{i=1}^{k}$ so that we can get the exact performance $L_{\mathsf{test}}(w^0_i)$ by using $w^0_i$. 
    % These samples are then split into training and validation sets. 
    \item \textbf{Estimating the optimal data mixture strategy $w$}. Using a non-linear second-order model $\widehat{L}_{\mathsf{test}}(w) := a + b^\top w + w^\top C w$, parameterized by $a \in \mathbb{R}$, $b \in \mathbb{R}^m$, and $C \in \mathbb{R}^{m \times m}$. This model captures both the independent (via $a$ and $b$) and correlated (via $C$) effects of different $m$ datasets on the performance function. The parameters $\{a, b, C\}$ are estimated by minimizing the empirical risk minimization (ERM) objective using the data generated according to the seed samples:
\begin{align}
    \min_{a, b, C} \frac{1}{k} \sum_{i=1}^k \mathcal{L} \left(\widehat{L}_{\mathsf{test}}(w^0_i), L_{\mathsf{test}}(w^0_i) \right),
    \label{eq:surrogate}
\end{align}
where $\mathcal{L}(\cdot, \cdot)$ is a fixed loss function (e.g., RMSE).  
\end{itemize}
Finally, computing the optimal weight $w^* = \arg\max_{w \in \Delta_m} \widehat{L}_{\mathsf{test}}(w)$ and using the corresponding training data distribution $\mu = \mu(w^\star)$ for RLVR post-training (i.e., solving \eqref{eq:RLVR}).

\section{Experiments}

% Define custom colors for the boxes
\definecolor{lightblue}{RGB}{230, 236, 250}
\definecolor{lightgreen}{RGB}{230, 242, 230}

% Define styles for response boxes
\newcommand{\responsebox}[2]{%
  \begin{tcolorbox}[
    colback=#1,
    colframe=#1,
    arc=5pt,
    width=0.85\textwidth,
    boxsep=5pt,
    left=10pt,
    right=10pt
  ]
    #2
  \end{tcolorbox}
}

\begin{wrapfigure}[12]{r}{.5\textwidth}
\vspace{-1cm}
\centering

\label{fig:w_wo_proc}

% Simple two-row table
\resizebox{.5\textwidth}{!}{
\begin{tabular}{c}
% First row - headers with more visible text
%~ & 
%\multicolumn{1}{c}{\colorbox{lightblue}{\textbf{w/o Processing}}} & 
%\multicolumn{1}{c}{\colorbox{lightgreen}{\textbf{ w/ Processing }}} \\[0.5cm]

 \includegraphics[width=0.7\textwidth]{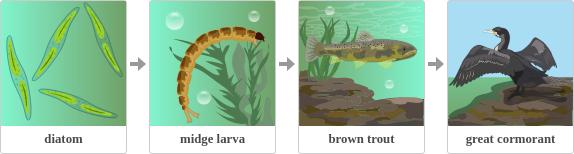}\\

% Second row - image and response boxes
%\multirow{2}{*}{\includegraphics[width=0.28\textwidth]{figures/w_wo_scienceqa_3.jpg}} & 
\responsebox{lightblue}{\textbf{Question:} In this food chain, the diatom is a producer. Why?

%Choices:\\
A. It eats another organism.
B. It makes its own food.
C. It eats a consumer.
\\
\textbf{Answer:} B 
}

\\

\responsebox{lightgreen}{\textbf{Question:} In this food chain, the diatom is a producer. Why?

%Choices:\\
A. It eats another organism.
B. It makes its own food.
C. It eats a consumer.\\
\textit{Output the thinking process in <think> </think> and final answer (option) in <answer> </answer> tags.}
\\
\textbf{Answer:} \textit{<think> In this food chain, the diatom is a producer because it makes its own food. The diatom uses carbon dioxide, water, and sunlight to make its own food. </think> <answer> B </answer>}
}
\\

\end{tabular}}
\caption{Demonstration of a General Question-Answer Pair \colorbox{lightgreen}{\textbf{With}} and \colorbox{lightblue}{\textbf{Without}} Reasoning.}
\end{wrapfigure}

We begin by describing the experimental setup, including dataset curation, reasoning mode, reward models, training and testing strategies, data sampling, and evaluation metrics in~\cref{sec:experimental_setting}. Subsequently,~\cref{sec:data mixture_implementation} explains the design of $3$ groups of data mixture strategies examined in our paper, i.e. seed, heuristic, and model-based. Finally,~\cref{sec:analysis} presents a comprehensive analysis of the results, demonstrating the importance of data mixture in Multimodal LLM RLVR and the efficacy of different data mixture strategies.
%\yl{Motivate the whole experiment section here!}

\subsection{Experimental Setup}
\label{sec:experimental_setting}

\paragraph{Data Curation}

To expose our model to a broad range of vision-language competencies---from scientific reasoning to fine-grained visual grounding---we assemble a diverse mixture of training corpora where each dataset's question-answer pairs are \emph{verifiable}: the ground-truth answer can be deterministically checked either by exact-string comparison (for multiple-choice letters or short free-form text) or by straightforward programmatic evaluation (for numeric responses).
This guarantees an unambiguous reward signal during RLVR Post-Training.

% \yl{add a section here explaining domain and distribution assumption?}

In addition to training datasets' in-distribution test splits (if exist), we adopt \emph{out-of-distribution} benchmarks that have been widely used to evaluate Multimodal LLMs~\cite{zhang2024lmmsevalrealitycheckevaluation}.  
Statistics for all training and evaluation sets appear in~\cref{tab:datasets}.

\begin{table}[t]
  \centering
  \caption{\textbf{Training and Testing datasets.} The top section shows \textbf{Training} datasets processed for RLVR Post-Training with Data Mixture. The bottom two sections show \textbf{Testing} datasets containing training datasets' in-distribution test sets (\textbf{In}) and common VLM test datasets (\textbf{Out}).
  }
  \label{tab:datasets}
  \resizebox{\textwidth}{!}{\begin{tabular}{cc|lllcr}
    \toprule
    \multicolumn{2}{c}{\textbf{Type}} &
    \textbf{Dataset} &
    \textbf{Domain} &
    \textbf{Answer Type} &
    \textbf{Rewards} &
    \textbf{\#\,samples} \\
    \midrule
    
    \multirow{5}{*}{\rotatebox[origin=c]{90}{\textbf{Training}}} & 
    & COCO~\cite{mscoco} & Object Recognition & 2D Bounding Box & IoU, Format & 5997 \\ 
    & & LISA-train~\cite{lai2023lisa} & Referring Expression  & 2D Bounding Box & IoU, Format & 1326 \\
    & & GeoQAV~\cite{li2025openr1multimodal} & Math VQA &  Multiple Choice & Acc, Format & 1969 \\ 
    & & SAT-train~\cite{ray2024satspatialaptitudetraining} & Spatial VQA & Natural Language & Acc, Format & 15000 \\
    & & ScienceQA-train~\cite{scienceqa} & Science VQA & Multiple Choice & Acc, Format & 6218 \\
    
    \cmidrule{2-7}
    
    \multirow{7}{*}{\rotatebox[origin=c]{90}{\textbf{Testing}}}
    & \multirow{3}{*}{\rotatebox[origin=c]{90}{\textbf{In}}}
    & LISA-test~\cite{lai2023lisa} & Referring Expression  & 2D Bounding Box & \cellcolor{gray!25} & 3397 \\
    & & SAT-test~\cite{ray2024satspatialaptitudetraining} & Spatial VQA & Natural Language & \cellcolor{gray!25} & 1928 \\
    & & ScienceQA-test~\cite{scienceqa} & Science VQA & Multiple Choice & \cellcolor{gray!25} & 2017 \\
    
    \cmidrule{3-4}\cmidrule{5-7}
    & \multirow{4}{*}{\rotatebox[origin=c]{90}{\textbf{Out}}} 
    & ChartQA~\cite{masry2022chartqa} & Chart VQA & Natural Language & \cellcolor{gray!25} & 2500 \\
    & &  InfoVQA~\cite{infovqa} & Infographics VQA & Natural Language & \cellcolor{gray!25} & 2801 \\
    & & MathVista~\cite{lu2024mathvista} & Math VQA & Multiple Choice & \cellcolor{gray!25} & 1000\\
    & & MMMU~\cite{yue2023mmmu} & General VQA & Multiple Choice & \cellcolor{gray!25} & 900 \\
    
    \bottomrule
  \end{tabular}
  }
\end{table}

\vspace{0.5em}
\noindent\textbf{Pre-Processing of Individual Datasets}
(\emph{some training datasets have corresponding testing datasets}):
\begin{enumerate}[noitemsep,topsep=0pt,leftmargin=1.5em]
    \item \textbf{COCO}~\cite{mscoco}: we adopt the full object-category subset used by \citet{liu2025visual}.
    \item \textbf{LISA}~\cite{lai2023lisa}: each image-bounding-box pair may be associated with multiple questions; we treat every question as an independent sample and shuffle the flattened list. Images whose longer side exceeds 640~pixels are resized while maintaining the aspect ratio. The same pipeline is applied to both the official train and test splits, yielding our LISA-train/test sets.
    \item \textbf{GeoQAV}~\cite{li2025openr1multimodal}: starting from the 8k-example math subset, we remove corrupted items by retaining only those whose answers match the expected multiple-choice regex pattern; oversized images are resized as in LISA.
    \item \textbf{SAT}~\cite{ray2024satspatialaptitudetraining}: we use the original train split for GRPO Post-Training and val split for evaluation.
    \item \textbf{ScienceQA}~\cite{scienceqa}: we keep only problems that reference an image. All answer options are concatenated with the original question into a single multiple-choice prompt, and the ground truth answer is converted to its corresponding letter label for style matching.
\end{enumerate}

\paragraph{Reasoning Mode}

\begin{wrapfigure}[15]{r}{.4\textwidth}
    \vspace{-.5cm}
    \centering
    
    %
    %\setlength{\fboxsep}{0pt}
    %\fbox{\includegraphics[trim={0 150pt 0 100pt},clip, width=\linewidth]{images/toby_sit-unstable.png}}
    \includegraphics[width=0.95\linewidth]{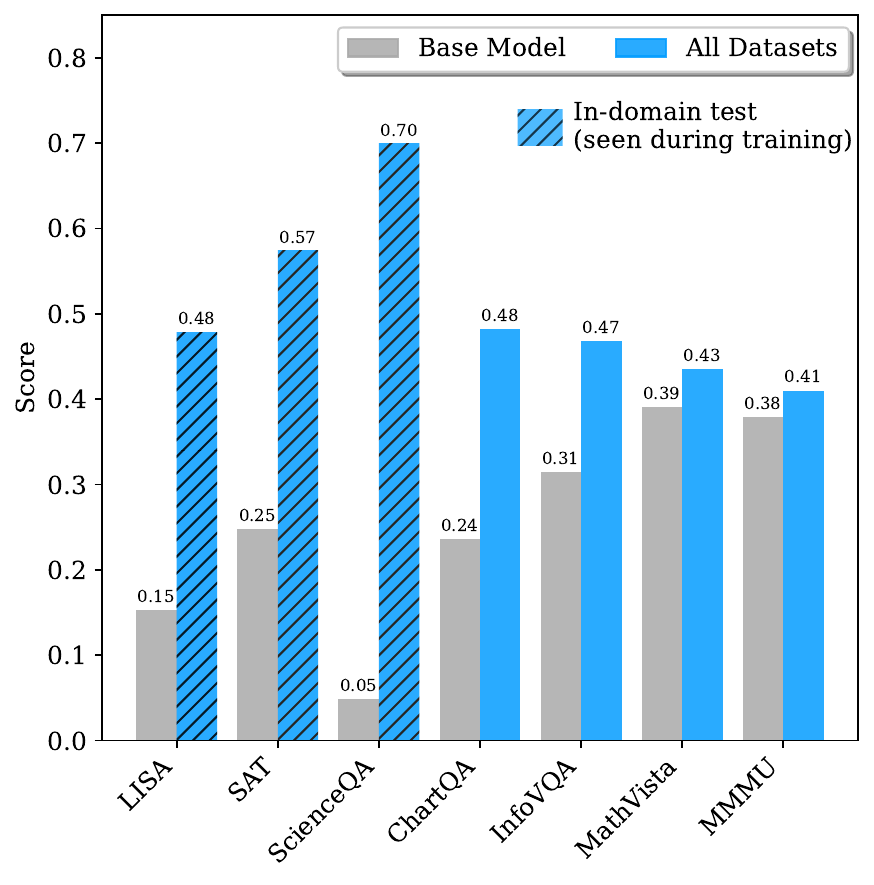}
    \vspace{-0.3cm}
    \captionof{figure}{Model Performance before / after GRPO training on \emph{All} data mixture.}
    \label{fig:base_vs_all}
\end{wrapfigure}

Rather than permitting the multimodal LLM to emit an answer immediately after observing the image-question pair, we place it in a two-step \emph{reasoning mode}:  
(1) generate a free-form chain-of-thought, and  
(2) commit to a concise final answer.  
We trigger this mode by appending a specialised reasoning prompt to every question.
\cref{fig:w_wo_proc} is an example.

\paragraph{Reward Models}
After the above steps, we extract the final answer via a regular-expression parse.  
\ding{182} \textbf{Format}: if the regex succeeds, this reward yields $1$; otherwise, \{Format, Acc, IoU\} are all set to zero.
\ding{183} Accuracy (\textbf{Acc}): a binary reward indicating an exact match.
\ding{184} Intersection-over-Union (\textbf{IoU}): for bounding boxes, a real-valued reward in $[0,1]$.
The specific combination of rewards applied to each training dataset is summarised in~\cref{tab:datasets}.

%\newpage
\paragraph{Train/Test Strategy}

To stay within $4\times$ NVIDIA A100 GPUS' budget while retaining adequate representational power, we select Qwen2-VL-2B-Instruct~\cite{Qwen2-VL} as the base for our GRPO experiments.  
To further improve efficiency, $1$ GPU runs a \texttt{vLLM}~\cite{vllm} server that handles all forward generation and periodically sync weight with other processes; the remaining GPUs execute policy update loop.
%\begin{itemize}[noitemsep,topsep=0pt,leftmargin=1.5em]
Each of $3$ training GPUs processes $2$ samples in parallel; $2$ gradient accumulation steps give an effective batch of $12$ trajectories per update. This provides the $\emph{k}=6$ generations needed for GRPO reward while maximizing throughput without exceeding memory limits.
The vision encoder is frozen; only the LLM-part parameters are updated.
We use AdamW optimizer with $\beta_1{=}0.9, \beta_2{=}0.999, \varepsilon{=}10^{-8}$.
The learning rate follows a linear schedule: $10\%$ warm-up to a peak $\eta_{\max}=1\times10^{-6}$, then linear decay to zero.
Training is performed in \texttt{bfloat16} with deterministic seed $42$.
Regarding GRPO-specific hyperparameters, we use KL coefficient $0.04$, reward weights $2.0$ for \textbf{Acc} and \textbf{IoU}, and $1.0$ for \textbf{Format}.
We train each experiment for $1$ epoch and report metrics after $2000$ steps of training. If a certain mixture training ends early, we use the last checkpoint for evaluation. We also compute metrics after $500$ and $1500$ steps of training for certain data mixture prediction strategies.

\paragraph{Training Data Sampling}

%As formalized in~\cref{sec:data mixture_modeling}, every dataset is regarded as a separate domain.  
Let ${w_1,\dots,w_m}$ denote the weights for the $m$ available training sets, with $\sum_{i=1}^{m}w_i=1$.  
For each training step, we perform a two-stage draw:
(1) sample a domain $(D_i)$ according to $(\Pr(D_i)=w_i)$;
(2) sample $1$ unseen example from the training split of $(D_i)$.
This design aims to force the model to see training samples proportional to the data mixture. If all training samples from a certain dataset have been seen, the training stops.
%This procedure realises importance sampling over domains while preserving unbiased sample selection within each dataset.
%\yl{update notation accordingly later on?}

\paragraph{Evaluation Metrics}

We compute a score in the range $[0,1]$ for every test dataset.  
For LISA-test, we use the mean intersection-over-union (IoU) metric between predicted and ground-truth boxes.  
For all others, we use the fraction of samples where the answer exactly matches the reference as the score.
To provide a concise summary of in/out-of-distribution, we additionally report $2$ aggregations.
\ding{182} \textbf{In-Score}: the weighted average over datasets that might be seen during training, and
\ding{183} \textbf{Out-Score}: the weighted average over datasets that were not used at train time.
Weights are proportional to the number of test samples in each dataset (see~\cref{tab:datasets}), ensuring that larger benchmarks contribute more to the overall score.

%\jht{Question I had at this point when reading as a naive reader: Looking at Table 1, why is MathVista considered to be out of distribution when it seems like it is similar to GeoQAV? e.g., has same categorization.}

%\jht{Then I went and looked at what is in MathVista - isn't ScienceQA in MathVista? ScienceQA is a training source.
%\url{https://huggingface.co/datasets/AI4Math/MathVista/blob/main/source.json}
%}

%

\subsection{Data Mixture Implementation}
\label{sec:data mixture_implementation}

%\yl{Check notation consistency later!}
%\jht{$\alpha$ should be $w$}

\paragraph{Seed data mixture}

We begin with a suite of \emph{seed} mixtures in which every participating dataset is equally sampled at each training step.  
Three variants are considered:

\begin{itemize}[noitemsep,topsep=0pt,leftmargin=1.5em]
    \item \textit{Single}: train on one dataset only ($\alpha_j = 1$ for the chosen dataset, $0$ otherwise).
    \item \textit{Exclude-One}: the mixtures uniformly over all but one dataset; the held-out set's weight is $0$.  
          %This variant reveals the marginal value of each source domain.
    \item \textit{All}: a mixture uniformly over the complete collection ($\alpha_i = 1/m$ for every dataset).
\end{itemize}

%These baselines serve as intuitive reference points and motivate the heuristic mixing strategies introduced in the next section.

\paragraph{Heuristic data mixture}

After the seed mixture strategies, 
we now explore \emph{heuristic} approaches that make assumptions about how each dataset impacts test performance. 
These data-driven strategies predict weights using empirical performance scores observed in our seed mixture experiments. 
We employ three heuristic methods:

%\yl{alpha: assume each training dataset would independently contribute to one test metric, }

\begin{itemize}[noitemsep,topsep=0pt,leftmargin=1.5em]
    \item \textit{Alpha-family}: These approaches assume independence of datasets, and track each dataset's contribution to performance with a hyperparameter $\alpha$ to balance In-Score and Out-Score priorities:
    \begin{itemize}
        \item \textit{A\textsubscript{in}} ($\alpha=1.0$)~/~\textit{A\textsubscript{out}} ($\alpha=0.0$): Preferred for In-Score~/~Out-Score
        \item \textit{A\textsubscript{bal}} ($\alpha=0.5$): Balanced weighting between In-Score and Out-Score
        %\item \textit{A\textsubscript{out}} ($\alpha=0.0$): Weights optimized for Out-Score
    \end{itemize}

    \item \textit{Collinearity-Aware Regression (Coli)}: Employs ridge regression and assumes that variance inflation factor (VIF) correction can account for statistical dependencies between datasets. %Coefficients are normalized after applying VIF correction and a non-negativity constraint, yielding weights that reflect each dataset's unique contribution to performance.

    \item \textit{Leave-One-Out Normalization (Norm)}: Assumes that the performance gap when a dataset is excluded compared to \emph{All} is related to its importance in the final mixture.
\end{itemize}

\paragraph{Model-based data mixture}

Moving beyond heuristic approaches, we introduce a Covariance Matrix Adaptation Evolution Strategy (CMA-ES) framework that fits a \emph{parametric-model} to approximate the mapping from data mixtures to performance scores.

After fitting empirical observations of both seed and heuristic mixture experiments, the parametric model enables the prediction of performance for any potential mixture without requiring additional empirical observations.
By sampling from regions centered on observed mixtures and ranking mixtures based on model estimation, we can discover mixtures of potential that heuristic methods could not reach, 
particularly when complex synergies exist between datasets.

%Unlike previous methods that rely on fixed assumptions about dataset contributions, 
%QuadSurface learns this relationship directly from empirical data by fitting a quadratic response surface that captures both individual dataset effects and their interactions. 

%This model-based approach offers two key advantages: first, it naturally improves as more experimental data becomes available, refining its predictions through continuous learning; 
%second, it enables performance prediction for any potential mixture without requiring additional experiments. 

\begin{wrapfigure}[18]{r}{.75\textwidth}
    \vspace{-.5cm}
    \centering
    
    %\setlength{\fboxsep}{0pt}
    %\fbox{\includegraphics[trim={0 150pt 0 100pt},clip, width=\linewidth]{images/toby_sit-unstable.png}}
    \includegraphics[width=\linewidth]{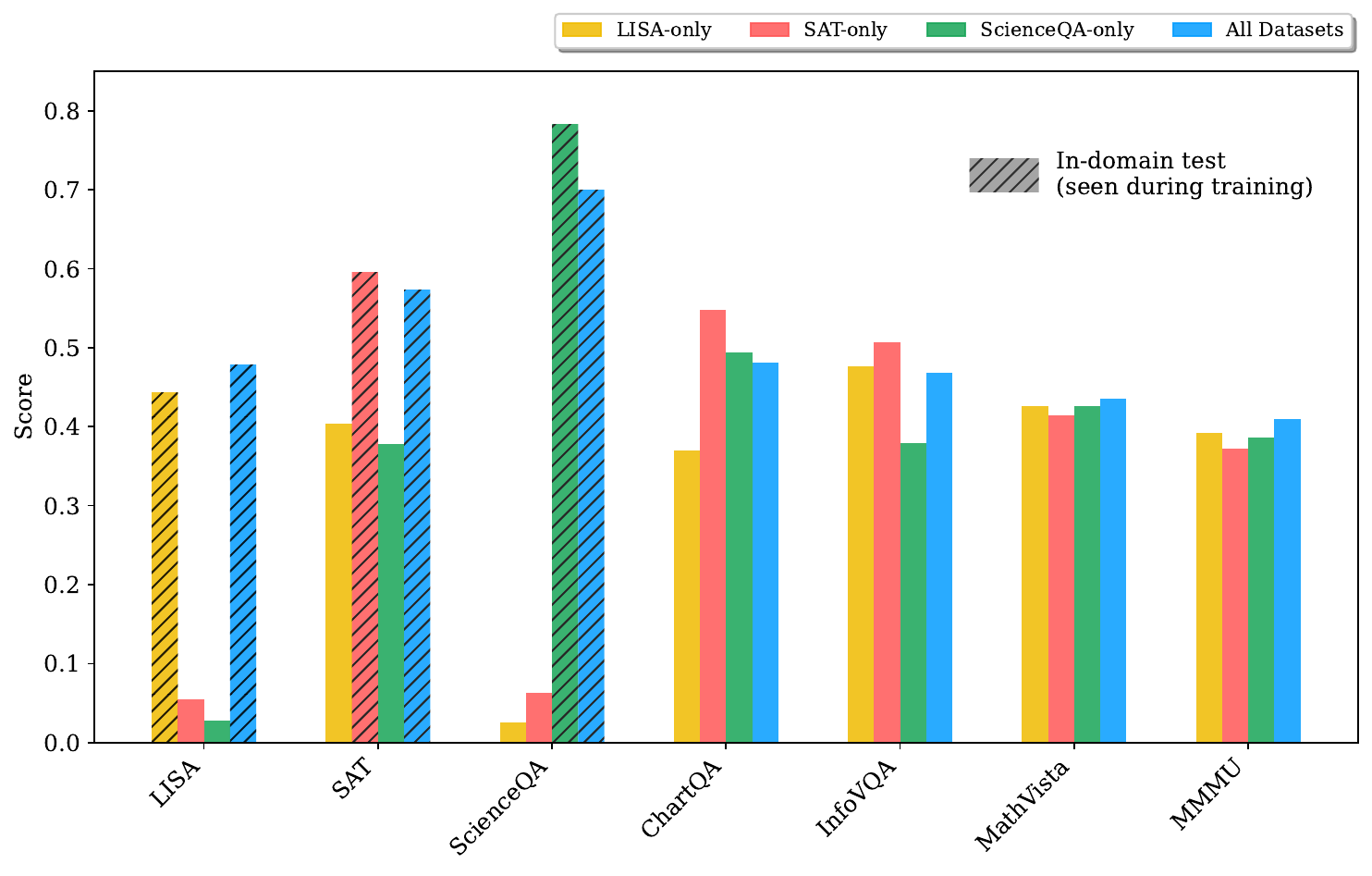}
    \vspace{-0.7cm}

    \captionof{figure}{Model Performance Comparison after GRPO training using \emph{All} data mixture and $3$ \emph{Single} data mixtures that have in-distribution test set.}
        
    \label{fig:single_vs_all}
\end{wrapfigure}

\subsection{Evaluation Results and Analysis}
\label{sec:analysis}

%Our experiments investigate the impact of data mixing for multimodal LLM post-training using Group Relative Policy Optimization (GRPO). 
We begin by showing the generalization potential of basic seed mixtures for multimodal LLM post-training using GRPO. Next, we explore the inherent complexities of combining diverse datasets. Lastly, we justify our parametric model choice for the model-based strategy, and compare the overall efficacy of seed, heuristic, and model-based-yielded data mixtures.% Result subset is in \cref{tab:main_performance}.

\begin{wrapfigure}[19]{r}{.75\textwidth}
    \centering
    %\vspace{-0.5cm}    
    \includegraphics[width=0.95\linewidth]{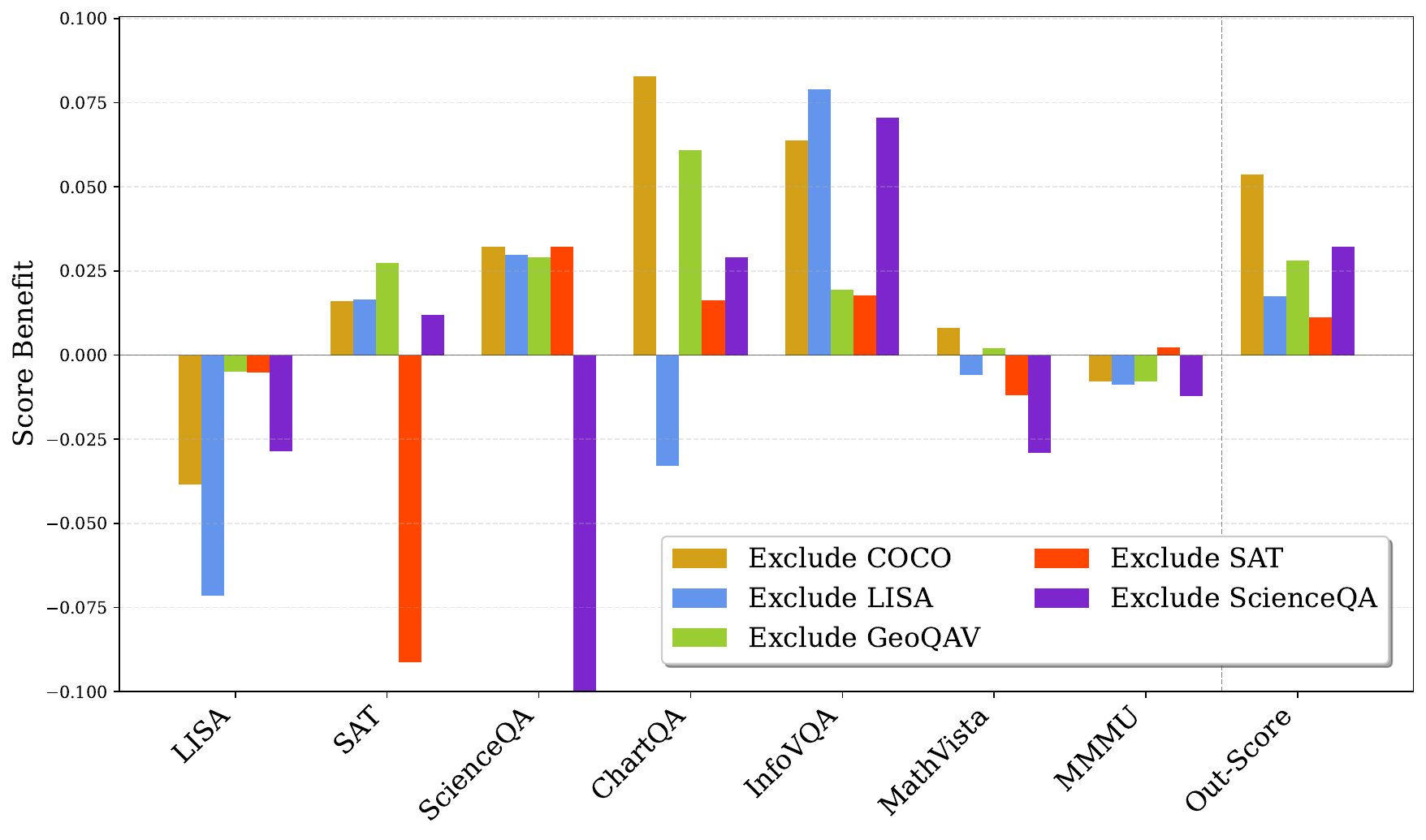}
    \vspace{-0.3cm}  
    \caption{Model Performance Comparison after GRPO training using \emph{All} data mixture and $5$ \emph{Exclude-One} data mixtures.}

    \label{fig:exclude_vs_all}
\end{wrapfigure}

\paragraph{Data Mixture Helps Generalization~(Figure~\ref{fig:base_vs_all}).}
Even a simple approach like seed data mixture can yield improvements over the base model, particularly in out-of-domain generalization. GRPO post-training with a seed mixture of all datasets~(\emph{All}) enhances model performance across in-domain test sets LISA, SAT, and ScienceQA, where the mixture leads to scores of $0.48$, $0.57$, and $0.70$ respectively, up from the base model's $0.15$, $0.25$, and $0.05$. More importantly, this trend of improvement extends to out-of-domain datasets including ChartQA ($0.24\rightarrow0.48$), InfoVQA ($0.31\rightarrow0.47$), MathVista ($0.39\rightarrow0.43$), and MMMU ($0.38\rightarrow0.41$). This indicates that mixture over diverse datasets, even without sophisticated weighting, can enhance both specialized and generalized capabilities.

\paragraph{Intricate Nature of Data Mixture (Figures~\ref{fig:single_vs_all} and \ref{fig:exclude_vs_all}).}
While the above seed mixture shows improvement on all test benchmarks, finding optimal data mixtures is far from straightforward.
First, naive inclusion of more data does not guarantee superior results. For example, \cref{fig:single_vs_all} shows that \emph{All}'s performance on ChartQA and InfoVQA is lower than $2$ single-dataset mixtures. 
Furthermore, \cref{fig:single_vs_all} illustrates that each dataset might have a distinct impact on model performance. For instance, LISA-only is beneficial for generalization on InfoVQA but hurts ScienceQA score, while ScienceQA-only yields the highest score on the same-domain test set but is less useful for InfoVQA.
When there's more than one data source, the interplay between them makes the situation more complicated. Figure~\ref{fig:exclude_vs_all} displays this complexity by comparing \emph{All} with \emph{Exclude-One} mixtures. The targeted removal of certain datasets might not negatively affect, or could even enhance, performance on particular tasks or overall. For instance, the exclusion of ScienceQA significantly lowers scores on the ScienceQA and ChartQA benchmarks, while its effects elsewhere are mixed. 
Additionally, excluding one dataset is consistently helpful for overall Out-Score, i.e., useful for out-domain generalization.
This variability underscores the need for more intelligent and adaptive data mixture strategies.

%This insight motivates the heuristic and model‑based mixers introduced in the following section, which aim to \emph{adaptively} allocate probability mass to each dataset rather than treating them as interchangeable bricks.

\begin{figure*}
    \centering
    \includegraphics[width=\textwidth]{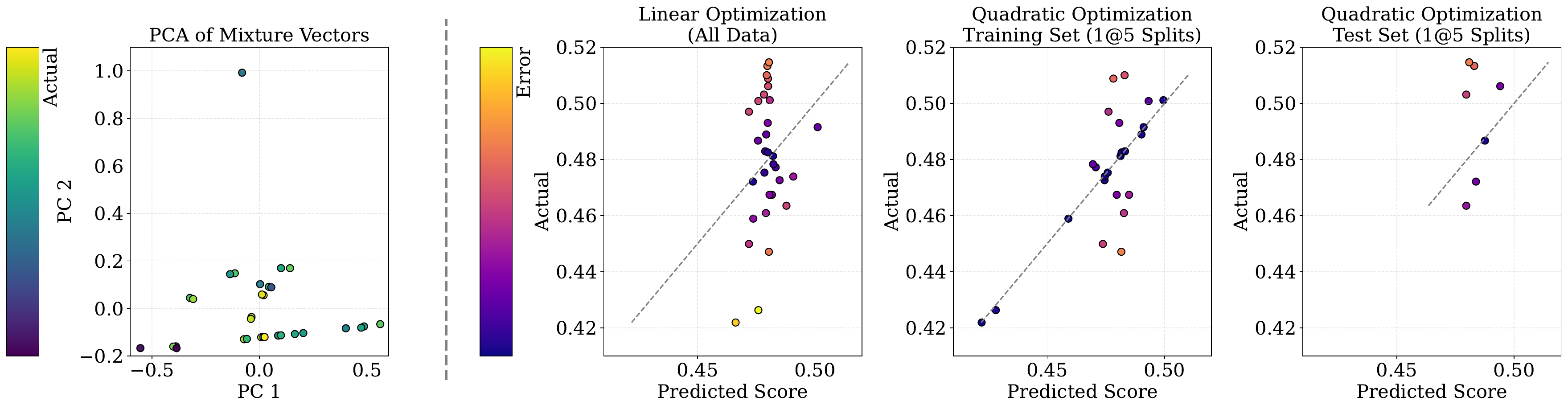}
    \caption{\textbf{Model-based mixture optimization.} \emph{Left:} PCA on mixture vectors reveals lack of linear separability, motivating nonlinear modeling. \emph{Middle-left:} Linear regression fails to fit Out-Score accurately, even with all data. \emph{Right:} Quadratic regression fitted on training folds could generalize better to held-out folds, capturing nonlinear interactions critical for mixture prediction.}
    \vspace{-.3cm}
    
    \label{fig:optim_fit}
\end{figure*}

\paragraph{Selection of an Effective Parametric data mixture Model (Figure~\ref{fig:optim_fit}).}
As the seed mixture shows promise but is likely suboptimal due to the aforementioned complexities, we investigated heuristic and model-based strategies to predict optimal data mixture. For model-based strategies, the choice of an appropriate mathematical model to approximate the function from mixture to Out-Score is crucial, as depicted in \cref{fig:optim_fit}. 
%Our experiments reveal that simpler modeling approaches are inadequate for capturing the complex, non-linear relationship between data mixture proportions and the resultant model performance. 
A linear (1-dimensional) optimization model, for example, fails to accurately map mixture to performance outcomes even when using all samples for fitting, as evidenced by the significant deviation between "Actual" and "Predicted Scores". Similarly, an analysis using Principal Component Analysis (PCA) on the mixture vectors suggests that the data points are not readily separable in a linearly reduced dimensional space, further pointing to underlying non-linearities.
In contrast, a quadratic optimization function is more adept. When employing cross-validation to pick top-$1$ out of $5$ shuffles, the quadratic model (rightmost panel) demonstrates improved ability to fit the training data and, critically, to generalize to unseen test sets. 
We hypothesize that the advantage of using a quadratic function over 1-dimensional linear function comes from the former's handling of the nuanced dataset interactions.
Now, we adopt the quadratic paradigm to guide the search for effective data mixture.

\begin{wrapfigure}[22]{r}{.4\textwidth}
    \vspace{-.4cm}
    
    \vspace{-.2cm}
    \centering
    %\setlength{\fboxsep}{0pt}
    %\fbox{\includegraphics[trim={0 150pt 0 100pt},clip, width=\linewidth]{images/toby_sit-unstable.png}}
    % trim={left bottom right top},clip]
    \includegraphics[trim={0 10pt 0 20pt},clip,width=0.95\linewidth]{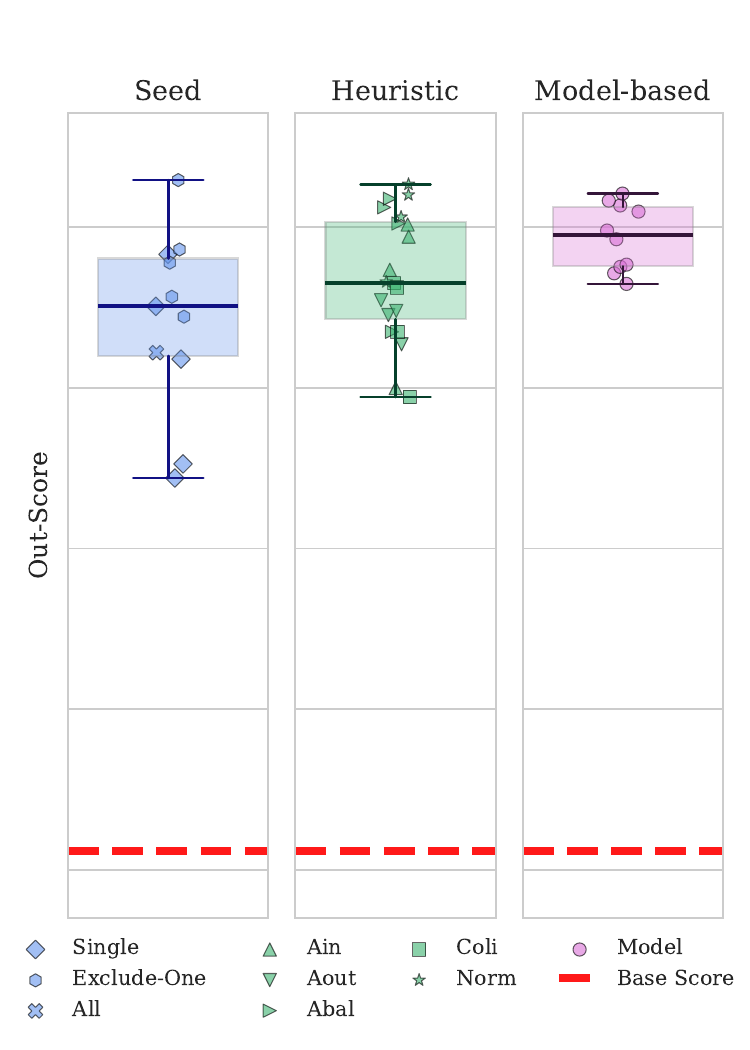}
    %\vspace{-0.65cm}
    \captionof{figure}{\textbf{Data Mixture Strategies' Comparison.} Heuristic~(\emph{middle}) and model-based~(\emph{right}) achieve higher minima and medians than Seed (\emph{middle}). Model-based further reduces variance.}
    
    \label{fig:min_max_var}
\end{wrapfigure}

\paragraph{Comparison of Mixture Strategies  (Figure~\ref{fig:min_max_var}).}
Our investigation of data mixture strategies is summarized in \cref{fig:min_max_var}, which presents an aggregated view of out-of-domain scores (Out-Score), grouped by strategy. This figure shows that, as groups, the predicted data mixtures of heuristic and model-based strategies both offer advantages over naive seed mixtures by having higher median and minimum scores. This reinforces the idea that thoughtful data mixing is the key to unlocking better generalization.

When comparing heuristic methods against our model-based approach (which leverages the aforementioned quadratic optimization), the model-based strategy achieves a higher median, a higher minimum, and a competitive maximum. Moreover, the model-based approach exhibits smaller variance compared to both seed and heuristic methods, which is a sign of not only an improvement in average performance but also a gain in reliability.

%In conclusion, while even basic seed mixtures can offer improvements, data mixture is a complex problem that requires where more sophisticated, model-based optimization using appropriate functional forms like quadratic models provides the most promising avenue for improving MLLM capacities, especially for out-of-domain generalization. \yl{ Mention A comprehensive table detailing all results is available in the supplementary material?}

%Setting @Yiqing Done

\section{Related Work}

\vspace{-5pt}
\paragraph{RL for aligning language models.}
Reinforcement Learning from Human Feedback (RLHF) guides LLMs with rewards learned from human‐labeled data, most visibly in InstructGPT~\citep{Ouyang2022instructgpt}.  Recent variants reduce human effort, e.g., Constitutional AI~\citep{Bai2022ConstitutionalAI}, RLAIF~\citep{Lee2024RLAIF}, and log-likelihood-based SimPO~\citep{Meng2024SimPO}, or replace value networks with groupwise comparisons (GRPO) and other verifiable-reward methods that markedly improve multi-step reasoning~\citep{Xiao2024DeepSeekMath,Chen2024LatentReasoning,Kazemnejad2024VinePPO}.

\vspace{-5pt}
\paragraph{R1-style multimodal reasoning.}
DeepSeek-Math/R1 show that GRPO can boost textual reasoning without an explicit reward model~\citep{deepseek-math}.  A fast-growing line of work ports this idea to vision–language models, demonstrating domain-specific gains in math, fine-grained recognition, counting, and spatial tasks (e.g., MAYE~\cite{ma2025maye}, Visual-RFT~\cite{liu2025visual}, R1-V~\cite{chen2025r1v}).  Follow-up studies refine rewards, curricula, and exploration strategies, achieving further stability and efficiency; yet how to allocate training across heterogeneous multimodal tasks remains open.

\vspace{-5pt}
\paragraph{Data-mixture strategies.}
Classic multilingual pretraining up-samples rare languages with temperature sampling~\citep{devlin2019bert}.  Proxy-model approaches such as DoReMi~\cite{xie2023doremi} and RegMix~\cite{liu2024regmix} learn mixture weights in two stages, while ODM~\cite{albalak2023efficient} updates them online.  Very recent evidence suggests that even simple heuristics rival these sophisticated schemes~\citep{held2025optimizing}.  We extend this discussion to vision–language reasoning, treating task-level sampling as a first-class design choice.

\noindent
Overall, prior studies show that (i) lightweight, verifiable rewards can reliably sharpen reasoning, and (ii) mixture policies critically shape downstream performance.  Our work unifies these threads by jointly optimising data mixture and GRPO‐style reinforcement for multimodal reasoning. For a full discussion, please refer to the supplementary materials Appendix~\ref{sec:appendix-related-work}.

\section{Conclusion}

We introduced \textbf{MoDoMoDo}, a framework that couples multi-domain data mixtures with reinforcement learning under verifiable rewards (RLVR) for multimodal large language models.
Leveraging five complementary image–text datasets equipped with rule-based reward functions, we train a lightweight \emph{quadratic surrogate} to predict post-training performance as a function of the mixture distribution.
This surrogate lets us identify an \emph{optimal} mix after only a small set of pilot runs. Models fine-tuned with the learned mixture surpass those trained on any individual dataset or on uniform combinations.

\paragraph{Limitations and Future Work.}
Our analysis is limited to image–text settings; extending MoDoMoDo to audio, video, and 3-D modalities, as well as to tasks where verifiable signals are sparse or noisy, remains open.
On the algorithmic side, exploring surrogate models that incorporate dataset similarity, curriculum schedules, or uncertainty estimates could further reduce pilot-run cost.
Finally, connecting our empirical findings to a unified multi-objective RL theory would deepen understanding of why mixture-optimized RLVR generalizes so well.

% We release all code, datasets, and trained checkpoints to spur further research on scalable, data-efficient post-training for multimodal foundation models.

\clearpage
{
\small
\bibliographystyle{abbrvnat}
\bibliography{main}
}

%%%%%%%%%%%%%%%%%%%%%%%%%%%%%%%%%%%%%%%%%%%%%%%%%%%%%%%%%%%%

\appendix
\newpage
%\section{Technical Appendices and Supplementary Material}
%Technical appendices with additional results, figures, graphs and proofs may be submitted with the paper submission before the full submission deadline (see above), or as a separate PDF in the ZIP file below before the supplementary material deadline. There is no page limit for the technical appendices.

\section{Related Work}\label{sec:appendix-related-work}
%\tzz{Can move related work to the second last section to make the writing flow better.}
% \tzz{If you don't have space, can cut it down to half page and mention a detailed literature review can be found in Appendix XXX}
\subsection{Reinforcement Learning for LLM}

Reinforcement Learning from Human Feedback (RLHF) has become a foundational technique for aligning LLM outputs with human preferences. Early implementations, such as InstructGPT~\citep{Ouyang2022instructgpt}, utilized Proximal Policy Optimization (PPO) guided by reward models trained on human-labeled data. Subsequent approaches like Constitutional AI~\citep{Bai2022ConstitutionalAI} introduced self-supervised objectives to reduce reliance on human feedback. To further minimize human involvement, Reinforcement Learning from AI Feedback (RLAIF)~\citep{Lee2024RLAIF} leverages AI-generated feedback. Simplifying the RLHF pipeline, SimPO~\citep{Meng2024SimPO} replaces explicit reward models with implicit rewards derived from the model's own log-likelihood, streamlining training while maintaining performance.

Building upon these foundations, recent research has focused on enhancing LLM reasoning capabilities through RL via verifiable rewards. DeepSeek's R1 models~\citep{Xiao2024DeepSeekMath, Liang2025DeepSeekR1} employ Group Relative Policy Optimization (GRPO), which eschews value networks in favor of comparing groups of outputs to reinforce correct reasoning patterns. Similarly, LaTRO~\citep{Chen2024LatentReasoning} frames reasoning as a latent-variable optimization problem, enabling self-improvement without external feedback. \cite{Havrilla2024} compare several RL algorithms (Expert Iteration, PPO, return-conditioned RL) using sparse correctness rewards or learned reward models, finding broadly similar gains in multi-step reasoning. VinePPO~\citep{Kazemnejad2024VinePPO} addresses credit assignment in multi-step reasoning by utilizing unbiased Monte Carlo returns, and Xiong et al.\ train models to iteratively generate, verify, and refine their answers using rule-based self-rewards \citep{Xiong2025SelfReward}.
\cite{Hu2025ORZ} show that a minimalistic PPO pipeline steadily increases response length and accuracy. On the systems side, VeRL \citep{Sheng2024HybridFlow} provides a flexible RL execution framework: it combines single- and multi-controller designs to support complex LLM training dataflows (PPO, GRPO, etc.) and reports 1.5$\times$-20$\times$ throughput improvements over prior systems.
Together, these advances demonstrate that relatively simple RL setups with rule-based correctness rewards can substantially enhance LLM reasoning when scaled effectively.

\subsection{R1-Style Multimodal Reasoning}
\label{sec:r1_reasoning}
  
Large language models (LLMs) are usually \emph{pre-trained} for next token generation and then \emph{post-trained} to follow instructions and reason.  The dominant post-training recipe combines supervised fine-tuning (SFT) with RLHF, where PPO is steered by a reward model built from human judgements~\cite{ouyang2022traininglanguagemodelsfollow}. Recently, DeepSeek-Math~\cite{deepseek-math} and DeepSeek-R1~\cite{deepseekai2025deepseekr1incentivizingreasoningcapability} skip the costly reward model training and replace PPO with \emph{Group-Relative Policy Optimisation}~(GRPO), achieving stronger reasoning at lower compute.  While DeepSeek-Math/R1 targeted language-only models, a wave of concurrent studies now adapts similar idea to multimodal large language models (MLLMs) to improve reasoning abilities~\cite{li2025surveystateartlarge,lin2025mindeyeslanguagereasoning}.

% ---------- middle part rewritten with connective openers ----------

\paragraph{Single-domain MLLM Reasoning}
The first wave of multimodal R1 work asks a focused question: \emph{can GRPO turn a general-purpose MLLM into a domain specialist?}  
\textbf{Math reasoning} is explored by MAYE~\cite{ma2025maye}, R1-OneVision~\cite{yang2025r1onevision}, and Multimodal-Open-R1~\cite{li2025openr1multimodal}.  
\textbf{Fine-grained/Open-vocabulary recognition} is addressed by Visual-RFT~\cite{liu2025visual} and VLM-R1~\cite{shen2025vlmr1}.  
R1-V~\cite{chen2025r1v} targets \textbf{visual counting and geometric reasoning}, and VisualThinker-R1-Zero~\cite{zhou2025r1zerosahamomentvisual} pioneers \textbf{spatial reasoning} without any SFT warm-up~\cite{ray2024satspatialaptitudetraining}.  
Collectively, these studies demonstrate that GRPO is \emph{task-portable}: by pairing a verifiable reward with a modest domain-specific corpus, one can reliably lift MLLM reasoning within that domain.

\paragraph{Reward engineering and algorithmic variants.}
Once a core stack exists, researchers turn to \emph{how} the policy is optimised, devising alternative rewards, KL schedules, and curricula.   OThink-MR1~\cite{liu2025othinkmr1stimulatingmultimodalgeneralized} introduces \textsc{GRPO-D}, an adaptive KL penalty that balances exploration and imitation for counting and geometry.  NoisyRollout~\cite{liu2025noisyrollout} injects image augmentations during roll-outs to strengthen mathematical reasoning, and ThinkLite-VL~\cite{wang2025sota} shows that a \emph{small but hard} subset can rival large-scale runs.  FAST~\cite{xiao2025fastslowthinkinglargevisionlanguage} further adapts the reasoning \emph{length} to task difficulty on-the-fly.

\paragraph{Stability, efficiency and exploration strategies.}
As reward designs grow more sophisticated, training can suffer from vanishing gradients and sample inefficiency, motivating techniques that explicitly stabilise and accelerate GRPO. \cite{chen2025suitabilityreinforcementfinetuningvisual} proposes normalized length reward to mitigate instability caused by compeletion length. VL-Rethinker~\cite{vl-rethinker} combats gradient issues via selective sample replay and forced self-reflection, whereas LUFFY~\cite{luffy} blends off-policy imitation with on-policy exploration using regularised importance sampling.

\paragraph{Empirical analyses and large-scale deployments.}
A complementary thread conducts systematic ablations or scales GRPO to production-grade MLLMs, clarifying its real-world impact.  VLAA-Thinking~\cite{chen2025sftrlearlyinvestigation} argues that SFT alone induces “pseudo-reasoning” and that mixed-reward GRPO restores genuine logical skill; Limit-of-RLVR~\cite{yue2025limit-of-rlvr} notes that GRPO mainly lifts top-1 rather than top-$k$ performance; and Perception-R1~\cite{yu2025perception} measures transfer to pure perception.  In practice, Skywork-R1V~\cite{skywork2025r1v} and Cosmos-Reason1~\cite{nvidia2025cosmosreason1physicalcommonsense} add a GRPO stage to large production models, while OpenVLThinker~\cite{deng2025openvlthinker} alternates SFT and GRPO over progressively harder datasets.  The recipe is already migrating to video—Video-R1~\cite{feng2025videor1}, SEED-Bench-R1~\cite{chen2025seedbenchr1} and VideoChat-R1—highlighting its modality-agnostic appeal.
% ---------- end of rewritten middle part ----------
To track progress, new visual puzzle suites probe the logical depth of MLLMs~\cite{song2025visualpuzzlesdecouplingmultimodalreasoning,leng2025crosswordbenchevaluatingreasoningcapabilities,xu2025visulogic}.

\paragraph{Open problem.}  
Despite the rapid proliferation of R1-style methods in MLLM Reasoning, \emph{data mixture strategies}—how to allocate training resources across heterogeneous reasoning tasks—remain unexplored.  Our work fills this gap and serves as a first step for future studies.

\subsection{Data Mixture for LLM}

Existing approaches to mixing heterogeneous data sources for pretraining have largely fallen into two categories: language-centric up-sampling and model-guided data mixture. In the multilingual setting, BERT~\cite{devlin2019bert} up-samples under-represented languages by raising sampling probabilities with a temperature hyperparameter, but this often causes the lowest-resource languages to be repeated excessively. To address this imbalance problem, UniMax~\cite{chung2023unimax} introduces a heuristic that dynamically adjusts language weights to achieve a fairer distribution across languages. In parallel, methods such as data mixing laws~\cite{ye2024data} learn analytic functions over sample mixtures to predict model performance on unseen combinations without requiring full training runs. DoReMi~\cite{xie2023doremi} and RegMix~\cite{liu2024regmix} pursue a similar two-stage strategy: a small proxy model is first trained -- via GroupDRO~\cite{sagawa2019distributionally} in DoReMi or by evaluating multiple mixture configurations in RegMix -- to identify optimal mixture weights, before resampling and training a full-scale model. Although effective, these proxy-based methods suffer from reduced efficiency due to multiple training passes, and the weights they learn often fail to generalize across different models. Online Data Mixing (ODM)~\cite{albalak2023efficient} builds on DoReMi by updating data weights during full-model training, but still relies on the same underlying minimax estimation.

More recent findings suggest that even simple heuristics, such as token-count proportions~\cite{held2025optimizing}, can outperform both manual and learned data mixture schemes, calling into question the necessity of complex weight estimation. Motivated by this insight, \cite{held2025optimizing} proposes UtiliMax and Model Estimated Data Utility (MEDU), which blend lightweight utility estimates with model-informed adjustments to strike a balance between simplicity and adaptivity. While the aforementioned techniques focus primarily on monolingual or multilingual text corpora drawn from sources like Wikipedia, books, web text, and code, our study shifts attention to the domains of vision-language reasoning tasks. In these multimodal settings, dataset relationships are governed not only by raw sample frequency but also by the semantic and reasoning objectives unique to each task, necessitating novel strategies for sampling and weighting across diverse VLM reasoning benchmarks.

% \yl{add a few data mix for VLM paper?}

\begin{figure}[htbp]
\centering
\caption{Comparison of Grounding Question-Answer Pairs \colorbox{lightgreen}{\textbf{With}} and \colorbox{lightblue}{\textbf{Without}} Reasoning.}
\label{fig:w_wo_proc_iou}

% Simple two-row table
\begin{tabular}{c}
% First row - headers with more visible text
%~ & 
%\multicolumn{1}{c}{\colorbox{lightblue}{\textbf{w/o Processing}}} & 
%\multicolumn{1}{c}{\colorbox{lightgreen}{\textbf{ w/ Processing }}} \\[0.5cm]

 \includegraphics[width=0.7\textwidth]{imgs/w_wo_lisa_1.png}\\

% Second row - image and response boxes
%\multirow{2}{*}{\includegraphics[width=0.28\textwidth]{figures/w_wo_scienceqa_3.jpg}} & 
\responsebox{lightblue}{\textbf{Question:} Output the bounding box in the image corresponding to the instruction: In a football match, which position primarily focuses on guarding the goal and stopping the opposing team from scoring?
\\
\textbf{Answer:} [422, 781, 464, 926] 
}

\\

\responsebox{lightgreen}{\textbf{Question:} Output the bounding box in the image corresponding to the instruction: In a football match, which position primarily focuses on guarding the goal and stopping the opposing team from scoring?\\
\textit{Output the thinking process in <think> </think> and final answer in <answer> </answer> tags.The output answer format should be as follows:
<think> ... </think> <answer>[{'Position': [x1, y1, x2, y2], 'Confidence': number}, ...]</answer>
Please strictly follow the format.}
\\
\textbf{Answer:} \textit{<think> ... </think> <answer>[{'Position': [422, 781, 464, 926], 'Confidence': 1}]</answer>
}
}\\

\end{tabular}
\end{figure}

% \yl{Need to talk about the speed difference between w. vllm and wo. vllm in suppl?}
% \yl{Need to talk about why decide to freeze vision model's evidences in suppl?}

\begin{table}[htbp]
\caption{Overall Performance Comparison.
Seed experiments use digits to denote training datasets: 
\textbf{1}: COCO, 
\textbf{2}: LISA,
\textbf{3}: GeoQAV,
\textbf{4}: SAT,
\textbf{5}: ScienceQA.}
\label{tab:overall_performance}
\resizebox{\textwidth}{!}{
\centering
%\begin{tabular}{llccc|cccc|>{\columncolor{gray!15}}c>{\columncolor{gray!15}}c}
\begin{tabular}{llccc|cccc|cc}
\toprule
\multicolumn{2}{c}{\multirow{2}{*}{\textbf{Experiment}}} & \multicolumn{3}{c|}{\textbf{In Tests}} & \multicolumn{4}{c|}{\textbf{Out Tests}} & \multirow{2}{*}{\textbf{In-Score}} & \multirow{2}{*}{\textbf{Out-Score}} \\
\cmidrule(lr){3-5} \cmidrule(lr){6-9}
\multicolumn{2}{c}{} & \textbf{LISA} & \textbf{SAT} & \textbf{ScienceQA} & \textbf{ChartQA} & \textbf{InfoVQA} & \textbf{MathVista} & \textbf{MMMU} & & \\
\midrule
\multicolumn{2}{c}{Base} & 0.1525 & 0.2479 & 0.0486 & 0.236 & 0.3144 & 0.391 & 0.3789 & 0.149 & 0.3059 \\
\midrule
\multirow{14}{*}{\rotatebox[origin=c]{90}{\textbf{Seed}}} 
& \multicolumn{10}{l}{\textit{Single}} \\
& \textbf{1} & 0.4015 & 0.5047 & 0.0258 & 0.41 & 0.5357 & 0.436 & 0.3811 & 0.3254 & 0.4589 \\
& \textbf{2} & 0.4433 & 0.4035 & 0.0253 & 0.3704 & 0.476 & 0.426 & 0.3922 & 0.318 & 0.4219 \\
& \textbf{3} & 0.0835 & 0.2547 & 0.4284 & 0.4936 & 0.5086 & 0.405 & 0.3989 & 0.2232 & 0.4753 \\
& \textbf{4} & 0.0551 & 0.5949 & 0.063 & 0.5476 & 0.5075 & 0.414 & 0.3722 & 0.199 & 0.4915 \\
& \textbf{5} & 0.0279 & 0.3786 & 0.7828 & 0.494 & 0.3787 & 0.426 & 0.3867 & 0.3274 & 0.4263 \\
& \multicolumn{9}{l}{\textit{Exclude-one}} \\
& \textbf{2345} & 0.4393 & 0.5897 & 0.7313 & 0.5644 & 0.5319 & 0.443 & 0.4022 & 0.559 & 0.5146 \\
& \textbf{1345} & 0.4064 & 0.5902 & 0.7288 & 0.4488 & 0.5471 & 0.429 & 0.4011 & 0.5432 & 0.4783 \\
& \textbf{1245} & 0.4728 & 0.6011 & 0.7283 & 0.5424 & 0.4875 & 0.437 & 0.4022 & 0.5767 & 0.4889 \\
& \textbf{1235} & 0.4727 & 0.4824 & 0.7313 & 0.498 & 0.4858 & 0.423 & 0.4122 & 0.5463 & 0.4721 \\
& \textbf{1234} & 0.4493 & 0.5856 & 0.4259 & 0.5108 & 0.5387 & 0.406 & 0.3978 & 0.4787 & 0.493 \\
& \multicolumn{9}{l}{\textit{All}} \\
& \textbf{12345} & 0.4778 & 0.5737 & 0.6991 & 0.4816 & 0.4681 & 0.435 & 0.41 & 0.5638 & 0.4609 \\
\midrule
\multirow{12}{*}{\rotatebox[origin=c]{90}{\textbf{Heuristic-based}}} 
& \multicolumn{9}{l}{\textit{Step-averaged}} \\
& $A_{in}$ & 0.4687 & 0.5928 & 0.7476 & 0.5524 & 0.5111 & 0.434 & 0.4 & 0.5779 & 0.5008 \\
& $A_{out}$ & 0.4616 & 0.5788 & 0.7179 & 0.488 & 0.5125 & 0.422 & 0.3989 & 0.5628 & 0.4772 \\
& $A_{bal}$ & 0.4755 & 0.5892 & 0.7338 & 0.552 & 0.5179 & 0.441 & 0.4144 & 0.5763 & 0.5061 \\
& \emph{Coli} & 0.4494 & 0.5773 & 0.7055 & 0.5304 & 0.4851 & 0.432 & 0.3978 & 0.5533 & 0.4825 \\
& \emph{Norm} & 0.4573 & 0.5664 & 0.705 & 0.564 & 0.5083 & 0.467 & 0.4133 & 0.554 & 0.51 \\
& \multicolumn{9}{l}{\textit{2000-step}} \\
& $A_{in}$ & 0.4415 & 0.5939 & 0.7581 & 0.522 & 0.5036 & 0.417 & 0.4133 & 0.5685 & 0.4867 \\
& $A_{out}$ & 0.4075 & 0.5799 & 0.71 & 0.4788 & 0.5089 & 0.429 & 0.3911 & 0.5359 & 0.4726 \\
& $A_{bal}$ & 0.4676 & 0.5877 & 0.7382 & 0.5548 & 0.5063 & 0.436 & 0.4078 & 0.5735 & 0.5011 \\
& \emph{Coli} & 0.4678 & 0.5783 & 0.6083 & 0.4076 & 0.5035 & 0.422 & 0.4089 & 0.5354 & 0.4471 \\
& \emph{Norm} & 0.4732 & 0.5762 & 0.7372 & 0.5764 & 0.5156 & 0.449 & 0.4022 & 0.5728 & 0.5133 \\
\midrule
\multirow{12}{*}{\rotatebox[origin=c]{90}{\textbf{Heuristic-based (No \textbf{1})}}} 
& \multicolumn{9}{l}{\textit{Step-averaged}} \\
& $A_{in}$ & 0.4498 & 0.402 & 0.7129 & 0.5456 & 0.5065 & 0.428 & 0.4089 & 0.5095 & 0.497 \\
& $A_{out}$ & 0.0873 & 0.5856 & 0.7015 & 0.4692 & 0.4932 & 0.421 & 0.4022 & 0.3869 & 0.4635 \\
& $A_{bal}$ & 0.4666 & 0.5892 & 0.7447 & 0.5572 & 0.5307 & 0.427 & 0.3967 & 0.5752 & 0.5088 \\
& \emph{Coli} & 0.448 & 0.5923 & 0.704 & 0.4836 & 0.5002 & 0.402 & 0.3933 & 0.5562 & 0.4674 \\
& \emph{Norm} & 0.4662 & 0.5794 & 0.711 & 0.5072 & 0.507 & 0.42 & 0.41 & 0.5632 & 0.4829 \\
& \multicolumn{9}{l}{\textit{2000-step}} \\
& $A_{in}$ & 0.4257 & 0.4414 & 0.7204 & 0.5096 & 0.4308 & 0.403 & 0.3956 & 0.5108 & 0.4499 \\
& $A_{out}$ & 0.0556 & 0.5731 & 0.6896 & 0.488 & 0.5053 & 0.415 & 0.4022 & 0.3657 & 0.4739 \\
& $A_{bal}$ & 0.4637 & 0.5529 & 0.7248 & 0.4904 & 0.4847 & 0.429 & 0.3922 & 0.5589 & 0.4674 \\
& \emph{Coli} & 0.4815 & 0.5897 & 0.7105 & 0.514 & 0.4978 & 0.42 & 0.4067 & 0.5728 & 0.4812 \\
& \emph{Norm} & 0.4065 & 0.5892 & 0.7169 & 0.5596 & 0.5158 & 0.424 & 0.3944 & 0.5398 & 0.5031 \\
\midrule
\multirow{10}{*}{\rotatebox[origin=c]{90}{\textbf{Model-based}}} 
& 001 & 0.2274 & 0.5871 & 0.7387 & 0.5196 & 0.5246 & 0.453 & 0.3911 & 0.4623 & 0.4962 \\
& 002 & 0.2674 & 0.5762 & 0.7288 & 0.5536 & 0.525 & 0.438 & 0.3956 & 0.4752 & 0.5067 \\
& 003 & 0.2225 & 0.5747 & 0.7427 & 0.514 & 0.5005 & 0.446 & 0.4044 & 0.4579 & 0.4856 \\
& 004 & 0.438 & 0.597 & 0.7154 & 0.5344 & 0.5016 & 0.424 & 0.3844 & 0.556 & 0.4876 \\
& 005 & 0.1219 & 0.6006 & 0.7313 & 0.4844 & 0.5256 & 0.414 & 0.4178 & 0.415 & 0.4823 \\
& 006 & 0.4293 & 0.5949 & 0.7402 & 0.5528 & 0.5126 & 0.427 & 0.3867 & 0.5582 & 0.4989 \\
& 007 & 0.2382 & 0.5669 & 0.7234 & 0.5412 & 0.5269 & 0.443 & 0.4033 & 0.4578 & 0.5048 \\
& 008 & 0.3843 & 0.5887 & 0.7214 & 0.516 & 0.5067 & 0.448 & 0.3989 & 0.5306 & 0.4883 \\
& 009 & 0.3955 & 0.5773 & 0.7343 & 0.5648 & 0.5258 & 0.434 & 0.3789 & 0.5363 & 0.5082 \\
& 010 & 0.3925 & 0.5742 & 0.7338 & 0.5668 & 0.5184 & 0.446 & 0.4 & 0.534 & 0.5104 \\
\bottomrule
\end{tabular}
}
\end{table}

% \yl{Not sure whether we need this anymore?
% Model merging @Yiqing Jiacheng 
% - Merge single model 
% - Merge weighted loss. 
% }

\begin{algorithm}
    \caption{Heuristic-Alpha: Balanced, In-optimized and Out-optimized}
    \label{alg:heuristic_alpha}
    \begin{algorithmic}[1]
    \Require
    \Statex $\mathcal{D} = \{1, ..., m\}$: Set of $m$ dataset indices
    \Statex $R$: Set of performance records from experiments
    \Statex $\alpha_{\text{single}} \in [0,1]$: Scaling factor for single datamix baselines
    \Statex $\alpha \in [0,1]$: Trade-off parameter balancing In and Out set performance
    \Ensure Dataset weights $\{w_i\}_{i \in \mathcal{D}}$ where $\sum_{i \in \mathcal{D}} w_i = 1$
    
    \State $S^{\text{in}} \gets [0]_{1..m}$ \Comment{Initialize In Score sums}
    \State $S^{\text{out}} \gets [0]_{1..m}$ \Comment{Initialize Out Score sums}
    
    \For{each record $\{s^{\text{in}}_r, s^{\text{out}}_r\} \in R$}
        \State Identify datasets $D_r \subseteq \mathcal{D}$ used in recipe $r$
        
        \If{$|D_r| = 1$} \Comment{Single-dataset}
            \State $s^{\text{in}}_r \gets \alpha_{\text{single}} \cdot s^{\text{in}}_r$
            \State $s^{\text{out}}_r \gets \alpha_{\text{single}} \cdot s^{\text{out}}_r$
        \EndIf
        
        \For{each dataset $i \in D_r$}
            \State $S^{\text{in}}[i] \gets S^{\text{in}}[i] + s^{\text{in}}_r$
            \State $S^{\text{out}}[i] \gets S^{\text{out}}[i] + s^{\text{out}}_r$
        \EndFor
    \EndFor

    \State \textbf{Min-max normalization:}
    \State $\hat{S}^{\text{in}} \gets \text{MinMax}(S^{\text{in}})$ \Comment{$\hat{S}^{\text{in}}_i = \frac{S^{\text{in}}_i - \min_j S^{\text{in}}_j}{\max_j S^{\text{in}}_j - \min_j S^{\text{in}}_j}$}
    \State $\hat{S}^{\text{out}} \gets \text{MinMax}(S^{\text{out}})$ \Comment{$\hat{S}^{\text{out}}_i = \frac{S^{\text{out}}_i - \min_j S^{\text{out}}_j}{\max_j S^{\text{out}}_j - \min_j S^{\text{out}}_j}$}
    
    \State \textbf{Combine scores with trade-off parameter:}
    \For{each dataset $i \in \mathcal{D}$}
        \State $C_i \gets \alpha \cdot \hat{S}^{\text{in}}[i] + (1 - \alpha) \cdot \hat{S}^{\text{out}}[i]$
    \EndFor
    
    \State \textbf{Normalize to obtain final weights:}
    \State $C_{\text{sum}} \gets \sum_{i \in \mathcal{D}} C_i$
    \For{each dataset $i \in \mathcal{D}$}
        \State $w_i \gets C_i / C_{\text{sum}}$
    \EndFor
    
    \State \Return $\{w_i\}_{i \in \mathcal{D}}$
    \end{algorithmic}
\end{algorithm}

\begin{algorithm}
    \caption{Heuristic-Colinearity; Variance Inflation Factor (VIF) quantifies multicollinearity.}
    \label{alg:heuristic_colinearity}
    \begin{algorithmic}[1]
    \Require
    \Statex $\mathcal{D} = \{1, ..., m\}$: Set of $m$ dataset indices
    \Statex $R$: Set of performance records from experiments
    \Statex $\lambda \in \mathbb{R}_+$: Ridge regularization parameter (default $10^{-3}$)
    \Ensure Dataset weights $\{w_i\}_{i \in \mathcal{D}}$ where $\sum_{i \in \mathcal{D}} w_i = 1$
    
    \State $X \gets [\,]$ \Comment{Initialize design matrix}
    \State $y \gets [\,]$ \Comment{Initialize target vector}
    
    \For{each record $\{s^{\text{in}}_r, s^{\text{out}}_r\} \in R$}
       
        \State Identify datasets $D_r \subseteq \mathcal{D}$ used in recipe $r$
        
        \State $x_r \gets [0]_{1..m}$ \Comment{Initialize feature vector for this record}
        \For{each dataset $i \in D_r$}
            \State $x_r[i] \gets 1$ \Comment{Set indicator to 1 for datasets used in recipe}
        \EndFor
        
        \State $X \gets X \cup \{x_r\}$ \Comment{Add feature vector to design matrix}
        \State $y \gets y \cup \{s^{\text{out}}_r\}$ \Comment{Add score to target vector}
    
    \EndFor
    
    \State \textbf{Fit ridge regression without intercept:}
    \State $\beta \gets \arg\min_{\beta} \|y - X\beta\|^2_2 + \lambda\|\beta\|^2_2$ \Comment{Ridge regression coefficients}
    
    \State \textbf{Compute Variance Inflation Factor (VIF) correction:}
    \State $I_m \gets \text{Identity matrix of size } m \times m$ \Comment{$m \times m$ identity matrix}

    \State $X^TX_{\text{reg}} \gets X^TX + \lambda \cdot I_m$ \Comment{Regularized Gram matrix}
    \State $(X^TX_{\text{reg}})^{-1} \gets \text{Inverse}(X^TX_{\text{reg}})$ \Comment{Inverse of Gram matrix}
    \State $\text{VIF} \gets \text{diag}((X^TX_{\text{reg}})^{-1})$ \Comment{Diagonal elements give VIF}
    
    \State \textbf{Adjust coefficients using VIF:}
    \For{each dataset $i \in \mathcal{D}$}
        \State $s_i \gets \beta_i / \text{VIF}_i$ \Comment{Adjust coefficient by its variance inflation}
    \EndFor
    
    \State \textbf{Apply non-negativity constraint:}
    \For{each dataset $i \in \mathcal{D}$}
        \State $s_i \gets \max(0, s_i)$ \Comment{Replace negative values with zero}
    \EndFor
    
    \State \textbf{Normalize to obtain final weights:}
    \State $s_{\text{sum}} \gets \sum_{i \in \mathcal{D}} s_i$
    \For{each dataset $i \in \mathcal{D}$}
        \State $w_i \gets s_i / s_{\text{sum}}$
    \EndFor
    
    \State \Return $\{w_i\}_{i \in \mathcal{D}}$
    \end{algorithmic}
\end{algorithm}

\begin{algorithm}
    \caption{Heuristic-Normalize: Leave-One-Out Dataset Weighting}
    \label{alg:heuristic_normalize}
    \begin{algorithmic}[1]
    \Require
    \Statex $\mathcal{D} = \{1, ..., m\}$: Set of $m$ dataset indices
    \Statex $R$: Set of performance records from experiments
    \Statex $C$: Set of leave-one-out experiments
    %\Statex $C = \{``1234", ``2345", ``1245", ``1235", ``1345"\}$: Set of leave-one-out combinations
    \Ensure Dataset weights $\{w_i\}_{i \in \mathcal{D}}$ where $\sum_{i \in \mathcal{D}} w_i = 1$
    
    \State $S_C \gets [\,]$ \Comment{Initialize scores for combinations}
    
    \For{each record $\{s^{\text{in}}_r, s^{\text{out}}_r\} \in R$}

        \State Extract combination ID $c_r$ from leading digits in $r$
        \State Extract out score $s^{\text{out}}_r$
        
        \If{$c_r \in C$} \Comment{If this is a leave-one-out combination}
            \State $S_C \gets S_C \cup \{(c_r, s^{\text{out}}_r)\}$ \Comment{Store combo and score}
        \EndIf
        
    \EndFor
    
    \State \textbf{Normalize out scores:}
    \State $s_{\max} \gets \max_{(c_r, s_r) \in S_C} s_r$ \Comment{Maximum out score}
    \State $s_{\min} \gets \min_{(c_r, s_r) \in S_C} s_r$ \Comment{Minimum out score}
    
    \For{each $(c_r, s_r) \in S_C$}
        \State $\hat{s}_r \gets \frac{s_r - s_{\min}}{s_{\max} - s_{\min}}$ \Comment{Min-max normalization}
        \State $s'_r \gets 0.2 - (0.1 \cdot \hat{s}_r)$ \Comment{Transform: higher score = lower weight}
    \EndFor
    
    \State \textbf{Derive dataset weights from leave-one-out scores:}
    \State $w_{\text{raw}} \gets [0]_{1..m}$ \Comment{Initialize raw weights}
    
    \For{$i \in \{1..m\}$}
        \State $w_{\text{raw}}[i] \gets s'_r \text{ where } c_r \text{ is the combination missing dataset } i$
    \EndFor
    
    \State \textbf{Normalize to obtain final weights:}
    \State $w_{\text{sum}} \gets \sum_{i=1}^{m} w_{\text{raw}}[i]$
    \For{$i \in \{1..m\}$}
        \State $w_i \gets w_{\text{raw}}[i] / w_{\text{sum}}$
    \EndFor
    
    \State \Return $\{w_i\}_{i \in \mathcal{D}}$
    \end{algorithmic}
\end{algorithm}

\begin{algorithm}
    \caption{QuadSurface: Quadratic Response Surface Optimization}
    \label{alg:quad_surface}
    \begin{algorithmic}[1]
    \Require
    \Statex $\mathcal{D} = \{1, ..., m\}$: Set of $m$ dataset indices
    \Statex $R$: Set of performance records from experiments
    \Statex $n_{\text{samples}}$: Number of candidate points to evaluate
    \Statex $k$: Number of top mixtures to return
    \Ensure Top $k$ dataset mixtures $\{w_i\}_{i \in \mathcal{D}}$ optimized for performance
    
    \State \textbf{Data preparation:}
    \State $X \gets [\,]$ \Comment{Initialize matrix of mixture weights}
    \State $y \gets [\,]$ \Comment{Initialize vector of performance scores}
    
    \For{each record $r \in R$}
        \State Extract weight vector $w_r$ and performance score $s_r$
        \State $X \gets X \cup \{w_r\}$
        \State $y \gets y \cup \{s_r\}$
    \EndFor
    
    \State \textbf{Cross-validation splits:}
    \State Perform $n_{\text{splits}}=5$ random train-test splits of $(X, y)$ \Comment{$80\%$ train, $20\%$ test}
    
    \For{each split $i$}
        \State $(X_{\text{train}}, y_{\text{train}}), (X_{\text{test}}, y_{\text{test}}) \gets \text{Split}_i(X, y)$
        
        \State \textbf{Fit quadratic response surface model:}
        \State $F_{\text{train}} \gets [1, X_{\text{train}}, X_{\text{train}} \otimes X_{\text{train}}]$ \Comment{Design matrix with quadratic terms}
        \State $\beta_i \gets \arg\min_{\beta} \|y_{\text{train}} - F_{\text{train}}\beta\|^2_2$ \Comment{Ordinary least squares}
        \State $(b_i, W_i, Q_i) \gets \text{Extract}(\beta_i)$ \Comment{Extract intercept, linear, and quadratic terms}
        
        \State \textbf{Evaluate model:}
        \State $\hat{y}_{\text{train}} \gets b_i + X_{\text{train}}W_i + (X_{\text{train}} \otimes X_{\text{train}})Q_i$
        \State $\hat{y}_{\text{test}} \gets b_i + X_{\text{test}}W_i + (X_{\text{test}} \otimes X_{\text{test}})Q_i$
        \State $R^2_{\text{train},i}, R^2_{\text{test},i} \gets \text{ComputeR}^2(\hat{y}_{\text{train}}, y_{\text{train}}, \hat{y}_{\text{test}}, y_{\text{test}})$
    \EndFor
    
    \State \textbf{Select best model:}
    \State $i^* \gets \arg\max_i R^2_{\text{test},i}$ \Comment{Choose model with best test performance}
    \State $(b^*, W^*, Q^*) \gets (b_{i^*}, W_{i^*}, Q_{i^*})$
    
    \State \textbf{Fit GMM to observed mixtures:}
    \State $\text{GMM} \gets \text{FitGaussianMixture}(X, n_{\text{components}}=1)$ \Comment{Fit single-component GMM}
    
    \State \textbf{Generate candidate mixtures via GMM:}
    
    \State $X_{\text{raw}} \gets \text{SampleFromGMM}(\text{GMM}, n_{\text{samples}})$
    \State $X_{\text{valid}} \gets \{x \in X_{\text{raw}} | x_i \geq 0 \text{ for all } i\}$ \Comment{Filter non-negative points}
    \State $X_{\text{candidates}} \gets \{x / \sum_i x_i | x \in X_{\text{valid}}\}$ \Comment{Normalize to sum to 1}

    \State \textbf{Predict performance for all candidates:}
    \State $\hat{y}_{\text{candidates}} \gets b^* + X_{\text{candidates}}W^* + (X_{\text{candidates}} \otimes X_{\text{candidates}})Q^*$
    
    \State \textbf{Select top-performing mixtures:}
    \State $\text{indices}_{\text{top}} \gets \text{Argsort}(\hat{y}_{\text{candidates}})[-k:]$ \Comment{Indices of top $k$ scores}
    \State $X_{\text{top}} \gets X_{\text{candidates}}[\text{indices}_{\text{top}}]$ \Comment{Top $k$ mixtures}
    
    \State \Return $X_{\text{top}}$
    \end{algorithmic}
\end{algorithm}

\end{document}